\documentclass[10pt,twocolumn,letterpaper]{article}

\usepackage{cvpr}              

\definecolor{cvprblue}{rgb}{0.21,0.49,0.74}
\usepackage[pagebackref,breaklinks,colorlinks,allcolors=cvprblue]{hyperref}
\usepackage{multirow}
\def\paperID{6061} 
\def\confName{CVPR}
\def\confYear{2025}

\title{Lifelong Knowledge Editing for Vision Language Models with 
\\Low-Rank Mixture-of-Experts}

\author{
\textbf{Qizhou Chen$^{1,3*}$, Chengyu Wang$^{2}\thanks{Q. Chen and C. Wang contributed equally to this work.}$, Dakan Wang$^{4}$, Taolin Zhang$^{5}$, Wangyue Li$^{1}$, Xiaofeng He$^{1}\thanks{Corresponding Author}$}\\
$^{1}$East China Normal University, Shanghai, China \\
$^{2}$Alibaba Cloud Computing, Hangzhou, China
$^{3}$Alibaba Group, Hangzhou, China\\
$^{4}$Exacity Inc., Shanghai, China
$^{5}$Hefei University of Technology, Hefei, China\\
{\tt\small 52265901009@stu.ecnu.edu.cn, chengyu.wcy@alibaba-inc.com, hexf@cs.ecnu.edu.cn}
}

\begin{document}
\maketitle

\begin{abstract}

Model editing aims to correct inaccurate knowledge, update outdated information, and incorporate new data into Large Language Models (LLMs) without the need for retraining. This task poses challenges in lifelong scenarios where edits must be continuously applied for real-world applications. While some editors demonstrate strong robustness for lifelong editing in pure LLMs, Vision LLMs (VLLMs), which incorporate an additional vision modality, are not directly adaptable to existing LLM editors. In this paper, we propose LiveEdit, a \underline{Li}felong \underline{v}ision language mod\underline{e}l \underline{Edit} to bridge the gap between lifelong LLM editing and VLLMs. We begin by training an editing expert generator to independently produce low-rank experts for each editing instance, with the goal of correcting the relevant responses of the VLLM. A hard filtering mechanism is developed to utilize visual semantic knowledge, thereby coarsely eliminating visually irrelevant experts for input queries during the inference stage of the post-edited model. Finally, to integrate visually relevant experts, we introduce a soft routing mechanism based on textual semantic relevance to achieve multi-expert fusion. For evaluation, we establish a benchmark for lifelong VLLM editing. Extensive experiments demonstrate that LiveEdit offers significant advantages in lifelong VLLM editing scenarios. Further experiments validate the rationality and effectiveness of each module design in LiveEdit. \footnote{The source code is available at \url{https://github.com/qizhou000/LiveEdit}.}

\end{abstract}

\section{Introduction}
Large language models (LLMs) have become key techniques for text generation in NLP~\cite{DBLP:journals/corr/abs-2302-13971,DBLP:journals/fi/RoumeliotisT23,DBLP:conf/iclr/ZengLDWL0YXZXTM23}. Benefiting from vision-language pre-training and pure LLMs, Vision-LLMs (VLLMs) are capable of generating text responses based on images and text~\cite{BLIP2,DBLP:conf/nips/Dai0LTZW0FH23,DBLP:journals/corr/abs-2308-12966,DBLP:journals/corr/abs-2309-15112}. 
However, outdated or erroneous built-in knowledge can undermine the value of these models.
To avoid the costly retraining of large-scale parameters, model editing aims to adapt models by adjusting a small number of parameters to update specific knowledge. This plays a critical role in areas such as privacy protection~\cite{DBLP:conf/acl/JangYYCLLS23,DBLP:journals/corr/abs-2309-11852}, detoxification~\cite{DBLP:conf/iclr/LiLCZLW0024,DBLP:conf/acl/Wang0XXDYZY0C24}, bias reduction~\cite{DBLP:conf/acl/YuJKYJ23,DBLP:conf/iclr/LimisiewiczMM24,DBLP:journals/corr/abs-2408-11843}, and hallucination correction~\cite{DBLP:conf/iclr/ChuangXLKGH24,DBLP:conf/acl/ZhangY024}.

\begin{figure*}[!t]
    \centering
    \includegraphics[width=1.\textwidth]{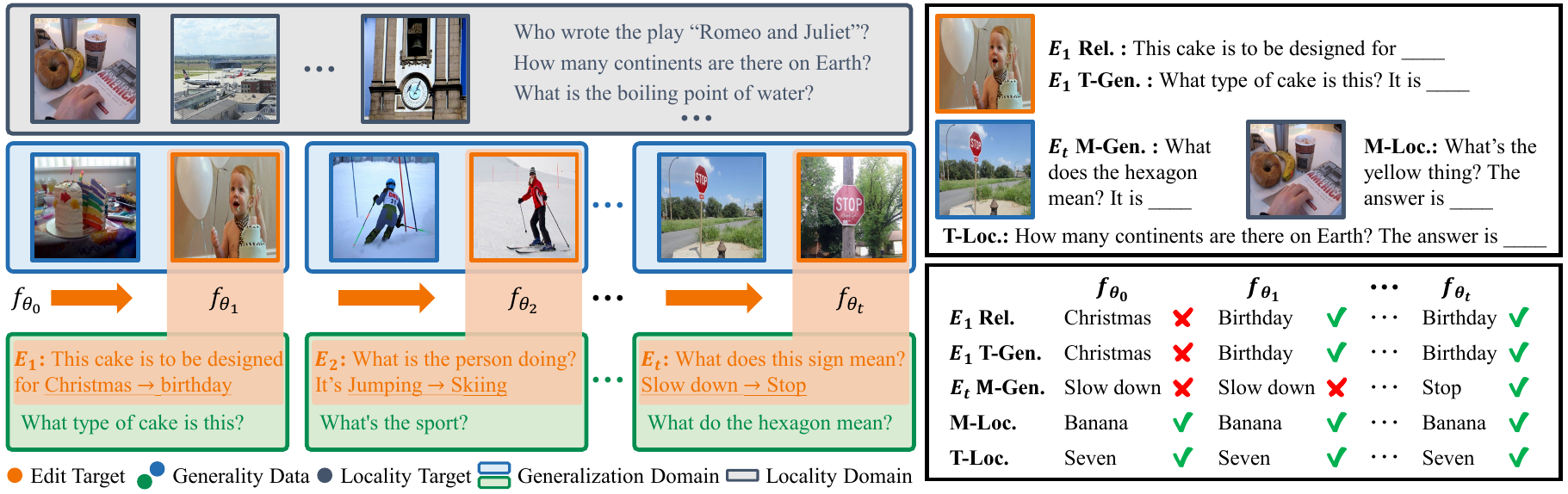}
\caption{Lifelong VLLM Editing. In this scenario, the edited VLLM is required to correctly respond to queries involving the edited data within the generalization domain, while maintaining consistent responses in locality domains. The top left shows test cases where Rel., T-Gen./M-Gen., T-Loc., and M-Loc. denote reliability, text/modal generality, and text locality, respectively. The bottom right illustrates the responses of an effectively edited VLLM across several editing timesteps.}
\label{fig_life_long_editing}
    \vspace{-1.0em}
\end{figure*}

Recent research on model editing techniques has primarily focused on pure LLMs~\cite{ROME,MEMIT,LTE,RECIPE}. 
However, the additional visual modality and the interactions between visual and textual modalities make these pure LLM editors less suitable. 
For example, LLM editors such as those in \cite{ROME,MEMIT,WILKE}, which are based on locate-then-edit methods, assume that the subject in the query is crucial for model reasoning.
These methods perform causal mediation analysis on input text queries containing the subject to identify linear layer weights critical to the LLM's reasoning. However, in vision-dominated tasks such as Visual Question Answering (VQA), where visual inputs often include substantial relevant information, attribution becomes more challenging. As a result, only limited work has explored how visual representations within VLLMs contribute to response generation and has proposed single-shot editing algorithms~\cite{MMEdit,DBLP:journals/corr/abs-2406-04236,DBLP:journals/corr/abs-2408-09916}.

In most LLM applications, single-shot model editing is insufficient to keep the model updated.
Thus, the concept of lifelong editing has emerged to address the continuous need for model updates~\cite{GRACE,WILKE,LTE,RECIPE}. In lifelong editing scenarios, some LLM editors have demonstrated strong performance. Retrieval-based methods, in particular, 
avoid directly editing the original model 
and apply on-the-fly edit retrieval and parameter fusion during inference~\cite{GRACE,LTE,RECIPE}. This approach offers greater robustness to a growing number of edits compared to editors that rely on permanent parameter modifications. 
For VLLMs, to the best of our knowledge, there is currently no related research on lifelong editing, as illustrated in Figure~\ref{fig_life_long_editing}. 
Visual modality is quite different from text modality since it typically contains more information and is more noisy.  
Therefore, the approaches working for LLMs cannot be directly applied to VLLMs.

In this paper, we introduce \emph{LiveEdit}, a novel framework designed for \underline{Li}felong \underline{v}ision language mod\underline{e}l \underline{Edit}ing, which bridges the gap between lifelong LLM editing and VLLMs. In this framework, we design a generative low-rank mixture-of-experts combined with hard and soft routing as a powerful VLLM editor. The two key techniques are outlined as follows:

\noindent\textbf{Generation of Low-Rank Experts:} 
Mixture-of-Experts (MoE) combines multiple "experts," each specializing in specific data patterns or sub-tasks~\cite{DBLP:journals/neco/JacobsJNH91,DBLP:conf/iclr/ShazeerMMDLHD17}. In VLLM editing, we treat VLLM’s adherence to a single edit sample as a sub-task, where each new edit sample corresponds to a low-rank expert that adjusts the model's response. MoE components are typically trained on sub-tasks, but directly fitting on individual edit samples results in poor generalization and inefficiency. To solve this, we propose an expert generator that creates low-rank experts for each new edit sample. The generator is trained to align VLLM with key edit metrics: reliability, textual/modal generality, and textual/modal locality~\cite{MMEdit}. The generated experts are stored in an expert repository. For new inputs, LiveEdit will select relevant experts from the repository and combines their adapted responses using hard and soft routing, as detailed below.

\noindent\textbf{Hard and Soft Routing:}
We propose a two-stage routing strategy for expert utilization during inference. Attribution in \cite{DBLP:journals/corr/abs-2408-09916} shows that VLLM processes prompts in early layers and extracts key visual features in later layers. In the first phase, we perform a text-to-vision interaction to extract key visual features and filter out noise. For each incoming sample, we compare its extracted features with those of edit samples, routing to visually relevant experts while filtering out visually irrelevant ones.
Since this hard routing only considers visual semantics, it may select multiple visually matched but text-irrelevant experts. In the second phase, we apply soft routing through multi-expert fusion, incorporating the semantic similarity between the input query and the edit text. By combining absolute and relative weights, relevant edit samples are assigned higher weights and irrelevant ones lower weights. This approach suppresses text-irrelevant experts and avoids redundant interactions.

In the experiments, the proposed LiveEdit framework is tested with 1, 10, 100, and 1,000 edits on the LLaVA-V1.5 (7B) \cite{llava}, MiniGPT-4 (7B) \cite{MiniGPT-4}, and BLIP2-OPT (2.7B) \cite{BLIP2} backbones across the E-VQA \cite{MMEdit}, E-IC \cite{MMEdit}, and VLKEB \cite{VLKEB} benchmark datasets. Comparisons with other strong editors demonstrated the superiority of our approach.

\section{Related Works}

\subsection{Vision Large Language Models}
Motivated by recent achievements of LLMs \cite{DBLP:journals/tist/ChangWWWYZCYWWYZCYYX24}, researchers have invested substantial effort in merging LLMs with vision models \cite{DBLP:conf/bigdataconf/WuGCWY23}.
VLLMs synchronize pre-trained image encoders, usually a Vision Transformer (ViT) \cite{DBLP:conf/iclr/DosovitskiyB0WZ21}, with an LLM decoder. Consequently, this configuration produces a model proficient in handling images alongside text inputs \cite{BLIP2,DBLP:conf/nips/Dai0LTZW0FH23,DBLP:journals/corr/abs-2308-12966,DBLP:journals/corr/abs-2306-16527,DBLP:journals/corr/abs-2309-15112}.
The training process for VLLMs typically unfolds in a two-phase approach. Initially, an alignment component, which may be a feed forward network \cite{llava} or more sophisticated structures such as a resampler \cite{DBLP:conf/nips/AlayracDLMBHLMM22,BLIP2}, is developed to bridge the image encoder with the LLM. 
This component is trained using pairs of images and their corresponding captions, effectively mapping image tokens onto the input space of the LLM. 
Subsequently, the focus shifts to broad-spectrum inference capabilities. The model is then refined through exposure to a diverse array of tasks, encompassing visual question-answering scenarios \cite{DBLP:conf/cvpr/GoyalKSBP17,DBLP:conf/cvpr/HudsonM19} and instruction-based interactions in both visual and textual contexts \cite{chiang2023vicuna,DBLP:conf/eccv/LiuDZLZZYWHLCL24}, thereby enriching its functional versatility. 
However, despite the broad application potential mentioned above, VLLMs still rely on meticulous fine-tuning and editing to ensure adaptability and accuracy across diverse scenarios.

\subsection{Model Editing}

\noindent\textbf{Model Editing for LLMs:}
We classify LLM editing into four categories.
(1) \textbf{Locate-Then-Edit} methods identify and modify specific model parameters related to target knowledge~\cite{DBLP:conf/acl/DaiDHSCW22}. ROME~\cite{ROME} uses causal mediation analysis for localization, while MEMIT~\cite{MEMIT} and WILKE~\cite{WILKE} extend it for multi-editing.
(2) \textbf{Meta Learning} methods employ a hyper-network to generate updated weights for edits ~\cite{KnowledgeEditor,MEND,DBLP:conf/iclr/TanZF24,DBLP:conf/acl/ZhangCL0HHXH24}.
(3) \textbf{Additional Parameters} methods introduce trainable parameters dedicated to edits while preserving original weights~\cite{T-Patcher, LEMOE}. Amon them, LEMOE~\cite{LEMOE} is based on MoE, but its greedy routing harms old experts' influence when integrating new ones.
(4) \textbf{Adding Extra Modules} methods store and retrieve edits via external memory mechanisms ~\cite{SERAC,GRACE,DBLP:conf/emnlp/MadaanTCY22,DBLP:conf/acl/WangHB24,LTE}.
In lifelong editing, the accumulation of shifts in the first two types hinders performance, while the latter two mitigate this by adding extra parameters and decoupling edits. However, in VLLM, extra modality and noise reduce efficacy.

\noindent\textbf{Model Editing for VLLMs:} Leveraging multimodal data for knowledge editing on VLLMs better resonates with practical contexts. Previously, the outlined methods were tailored for LLMs, operating solely on single-modal data. 
Yet, when it comes to knowledge editing on VLLMs, employing multimodal data offers a closer approximation to real-life settings. 
In the literature, MMEdit~\cite{MMEdit} and  VLKEB \cite{VLKEB} contribute novel datasets designed specifically for multimodal knowledge editing tasks. 
\cite{DBLP:journals/corr/abs-2408-09916} use attribution analysis to explore how VLLMs extract key information from visual representations to generate responses, leading to the design of a single-step VLLM editing technique.

\subsection{Mixture-of-Experts (MoE)}
The MoE technique \cite{DBLP:journals/neco/JacobsJNH91,DBLP:conf/iclr/ShazeerMMDLHD17} decomposes complex tasks into simpler ones, using dedicated models called experts. Recently, MoE layers have been integrated into transformer architectures. For instance, GShard \cite{DBLP:conf/iclr/LepikhinLXCFHKS21} utilized MoE in transformer, achieving significant improvements in machine translation for 100 languages. Switch Transformers \cite{DBLP:journals/jmlr/FedusZS22} further scaled language models with a trillion parameters through efficient MoE designs.
However, naive MoE training may cause load imbalance, wherein a few experts are overused while others are underutilized. To combat this, various strategies such as the BASE layer \cite{DBLP:conf/icml/LewisBDGZ21}, HASH layer \cite{DBLP:conf/nips/RollerSSW21}, and Expert Choice \cite{DBLP:conf/nips/ZhouLLDHZDCLL22} have been developed to optimize MoE models' capacity. Recent efforts focus on training a decoder-only MoE model with a modified UL2 objective \cite{DBLP:conf/icml/XueZFNZZ024}. Notably, Mixtral \cite{mistral2023mixtral} enhances decision-making by employing token-choice routing to select two out of eight experts, improving overall performance.
Our work approaches the MoE structure from a different angle, treating each knowledge update as a mini-task and leveraging low-rank experts to store knowledge for VLLM editing.

\begin{figure*}[!t]
    \centering
    \includegraphics[width=0.9\textwidth]{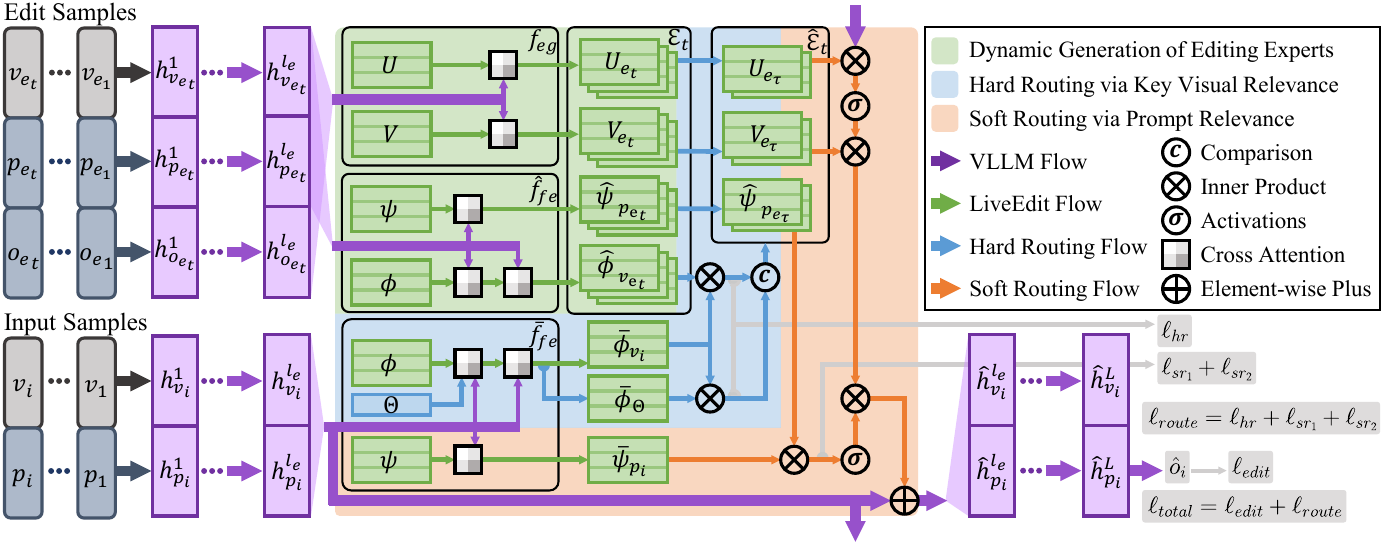}
    \vspace{-0.8em}
\caption{
Illustration of the LiveEdit framework. 
The upper part illustrates the editing process of LiveEdit. 
At time step $t$, the representation of an edit sample $(v_{e_t},p_{e_t},o_{e_t})$ at layer $l_e$ serves as an editing signal to generate the editing expert $(U_{e_t}, V_{e_t})$ via $f_{eg}$ and routing features $(\hat{\phi}_{v_{e_t}},\hat{\psi}_{p_{e_t}})$ via $\hat{f}_{fe}$. Both are then added to the expert repository $\mathcal{E}_{t}$.
The lower part shows the VLLM inference process with LiveEdit, where $\bar{f}_{fe}$ extracts input sample features at layer $l_e$ to route editing experts, which then adapt the representation.
}
    \label{fig_model}
    \vspace{-1.em}
\end{figure*}
 
\section{Preliminaries and Task Definition}
We formally define the VLLM editing task and its lifelong extension. Next, we introduce the evaluation criteria.

A VLLM $f_\theta:\mathcal{V}\times\mathcal{P}\rightarrow \mathcal{O}$ can be considered a function that maps an image-prompt pair $(v, p)$ to a textual output $o=f_\theta(v, p)$. Given an edit sample $(v_e, p_e, o_e)$, where $f_\theta(v_e, p_e) \neq o_e$, a VLLM editor $\text{ME}:\mathcal{F}\times \mathcal{V}\times\mathcal{P}\times \mathcal{O} \rightarrow \mathcal{F}$ produces an updated VLLM $f_{\theta'}=\text{ME}(f_\theta, v_e, p_e, o_e)$. 
Starting from an initial VLLM $f_{\theta_0}$, ME iteratively applies edits as new editing requirements arise in a lifelong context:
\begin{gather*}
    f_{\theta_t} = \text{ME}(f_{\theta_{t-1}}, v_{e_t}, p_{e_t}, o_{e_t}), t = 1,2,3,...
\end{gather*}
At any timestep $t$, an effective $\text{ME}$ should ensure that $f_{\theta_t}$ satisfies the following three criteria, as outlined in \cite{MMEdit}:

\noindent\textbf{Reliability} measures the accuracy of the modified model's responses on edited samples: 
\begin{gather*} 
\mathbb{E}_{(v_e, p_e, o_e) \sim \{(v_{e_\tau}, p_{e_\tau}, o_{e_\tau})\}_{\tau=1}^t} \mathbb{I}\{f_{\theta_t}(v_e, p_e) = o_e\} 
\end{gather*} 
where $\mathbb{I}$ is the indicator function that evaluates to 1 when the condition is true.

\noindent\textbf{Generality} requires $f_{\theta_t}$ can also adapt to relevant variations (e.g., rephrased prompts) in the edited samples, including modal and text generality: 
\begin{align*}
\mathbb{E}_{(v_e, p_e, o_e) \sim \{(v_{e_\tau}, p_{e_\tau}, o_{e_\tau})\}_{\tau=1}^t} \mathbb{E}_{v_g\sim \mathcal{G}(v_e)} 
\mathbb{I}\left\{f_{\theta_t}(v_g, p_e) = o_e\right\}\\
\mathbb{E}_{(v_e, p_e, o_e) \sim \{(v_{e_\tau}, p_{e_\tau}, o_{e_\tau})\}_{\tau=1}^t} \mathbb{E}_{p_g\sim \mathcal{G}(p_e)} 
\mathbb{I}\left\{f_{\theta_t}(v_e, p_g) = o_e\right\}
\end{align*}
where $\mathcal{G}(\cdot)$ represents the relevant neighbors.

\noindent\textbf{Locality} requires $f_{\theta_t}$ remains consistent with $f_{\theta_0}$ for samples unrelated to edits, including modal and text locality:
\begin{gather*}
\mathbb{E}_{(v_e, p_e, o_e) \sim \{(v_{e_\tau}, p_{e_\tau}, o_{e_\tau})\}_{\tau=1}^t} \mathbb{E}_{(v_l, p_l)\sim \mathcal{L}(v_e,p_e)} 
\mathbb{I}_l(v_l,p_l)\\
\mathbb{E}_{(v_e, p_e, o_e) \sim \{(v_{e_\tau}, p_{e_\tau}, o_{e_\tau})\}_{\tau=1}^t} 
\mathbb{E}_{p_l\sim \mathcal{L}(p_e)} 
\mathbb{I}_l(\emptyset, p_l)\\
\text{s.t.}\;\;\mathbb{I}_l(v, p) = \mathbb{I}\left\{f_{\theta_t}(v, p) = f_{\theta_0}(v, p)\right\}
\end{gather*}
where $\mathcal{L}(\cdot)$ represents the irrelevant samples.

\section{The Proposed LiveEdit Framework}

In this section, we formally introduce the LiveEdit framework, with the overall architecture shown in Figure \ref{fig_model}. 
First, we explain how to generate corresponding experts for edit samples and how to extract the semantic features of the editing prompt. We then describe their use in extracting key visual features from the visual representation, thereby maintaining the expert repository. 
Next, we describe how LiveEdit routes experts during a single inference of the VLLM to instantly adjust its response. 
Finally, we elaborate on the overall training process of the LiveEdit model.

\subsection{Construction and Update of Expert Repository}
Since the LLM transformer is the primary module for semantic understanding and response generation, in this work, we consider inserting the MoE editor between the layers of the transformer for editing. Previous work \cite{DBLP:journals/corr/abs-2408-09916} on representation attribution in VLLM editing has shown that the latter layers in the model leverage the prompt semantics to extract relevant visual information to generate responses. Following their findings, we deploy our editor in a high-contribution layer $l_e$ within the transformer. 
The expert repository is initially set to $\mathcal{E}_0 = \{\}$ and is updated from $\mathcal{E}_{t-1}$ to $\mathcal{E}_t$ at timestep $t$ as a new editing sample $(v_{e_t}, p_{e_t}, o_{e_t})$ is input into the model.

Specifically, given a VLLM $f_\theta$, the image and text of the edit sample are converted into embeddings, concatenated, and fed into the transformer. Let $h^{l_e} \in \mathbb{R}^{N \times d}$ represent the intermediate output at layer $l_e$, where $N$ and $d$ correspond to the sequence length and the intermediate dimension, respectively. Let $h_{v_{e_t}}^{l_{e}}$, $h_{p_{e_t}}^{l_e}$, $h_{o_{e_t}}^{l_e}$ denote the respective representations of $v_{e_t}$, $p_{e_t}$, and $o_{e_t}$. We define $f_{eg}(\cdot)$ to extract the editing signal and generate the expert:
\begin{equation}
\begin{aligned}
(U_{e_t}, V_{e_t}) &= f_{eg}(h_{v_{e_t}}^{l_e} \oplus h_{p_{e_t}}^{l_e} \oplus h_{o_{e_t}}^{l_e}),\\
\text{s. t.}\;\; f_{eg}(h) &= \left(\text{CA}_U(U, h), \text{CA}_V(V, h) \right)\label{eq_expert_generation}
\end{aligned}
\end{equation}
where $U \in \mathbb{R}^{r \times d_m}$ and $V \in \mathbb{R}^{r \times d_m}$ are two trainable matrices.
$\oplus$ denotes concatenation. $r$ and $d_m$ are hyper-parameters representing the number of ranks and the module dimension, respectively. The cross attention $\text{CA}(\cdot)$ is formulated as:
\begin{gather}
\text{CA}(x, y) = \delta\left(xW_q\left(yW_k\right)^T\right)\cdot yW_v
\end{gather}
where $\delta$ denotes softmax, and $W_q, W_k, W_v$ are matrices that map inputs into query, key, and value spaces, respectively.

To perform both hard and soft routing for experts, we respectively extract key visual feature $\hat{\phi}_{v_{e_t}}$ using prompt semantics and the pure prompt feature $\hat{\psi}_{p_{e_t}}$
through a feature extractor $\hat{f}_{fe}(\cdot)$:
\begin{equation}
\begin{aligned}
(\hat{\phi}_{v_{e_t}}, \hat{\psi}_{p_{e_t}}) &= \hat{f}_{fe}(h_{v_{e_t}}^{l_e}, h_{p_{e_t}}^{l_e})\label{eq_feature_extraction}\\
\text{s.t.}\quad \hat{f}_{fe}(h_v,h_p) &=
\left(
\text{CA}_{\phi2}\left(\text{CA}_{\phi1}\left(\phi, h_p\right),h_v\right),
\text{CA}_{\psi}\left(\psi, h_p\right)
\right)
\end{aligned}
\end{equation}
where $\phi \in \mathbb{R}^{1\times k  d_m}$ and $\psi \in \mathbb{R}^{1 \times k d_m}$ are trainable feature extraction vectors, and $k$ controls the dimension of the vectors. The extracted features $\hat{\phi}_{v_{e_t}}$ and $\hat{\psi}_{p_{e_t}}$ have the same shape as $\phi$ and $\psi$. Finally, the expert repository is updated by inserting the group of experts and the routing features as $\mathcal{E}_{t} = \mathcal{E}_{t-1}\cup\{(U_{e_t},V_{e_t},\hat{\phi}_{v_{e_t}}, \hat{\psi}_{p_{e_t}})\}$.

\subsection{Expert Routing and Editing on the Fly}
Given an input image-prompt pair $(v_i, p_i)$, let its output at the $l_e$-th layer be $h^{l_e}$, where $h_{v_i}^{l_e}$ and $h_{p_i}^{l_e}$ denote the components corresponding to $v_i$ and $p_i$, respectively. We use an additional feature extraction function $\bar{f}_{fe}$  (defined in consistency within Eq.\ref{eq_feature_extraction}, but taking different inputs) to extract routing features from the input:
\begin{equation}
\begin{aligned}
(\bar{\phi}_{v_i}, \bar{\psi}_{p_i}) = \bar{f}_{fe}(h_{v_i}^{l_e}, h_{p_i}^{l_e}).
\end{aligned}
\end{equation}
Given $\mathcal{E}_t=\{(U_{e_\tau}, V_{e_\tau}, \hat{\phi}_{v_{e_\tau}}, \hat{\psi}_{p_{e_\tau}})\}_{\tau = 1}^t$, we filter for experts that are highly relevant to the input sample's visual content by calculating the similarity between the key visual features of the input and the edit samples:
\begin{equation}
\begin{aligned}
\hat{\mathcal{E}} = \left\{(U_{e_\tau}, V_{e_\tau}, \hat{\psi}_{p_{e_\tau}}) \mid 
\bar{\phi}_{v_i} \hat{\phi}_{v_{e_\tau}}^T > \bar{\phi}_{v_i} \bar{\phi}_{\Theta}^T, \; \tau = 1,...,t \right\}, \\
\text{s.t.} \;\; (\bar{\phi}_{\Theta}, \_) = \bar{f}_{fe}(\Theta, h_{p_i}^{l_e})
\label{eq_soft_routing_and_visual_sentinel}
\end{aligned}
\end{equation}
where a trainable vision sentinel $\Theta \in \mathbb{R}^{N_v \times d}$ is set to dynamically determine the filtering threshold, following \cite{RECIPE}, which effectively avoids the bias caused by manually set thresholds. $N_v$ is the vision token count of $f_\theta$. 
Intuitively, if the input sample is more visually similar to the edit sample than the visual sentinel, then this edit sample should not be selected.

Although the above process effectively selects visually relevant editing experts, some results may still have low prompt semantic relevance. We further use the similarity between prompt features to achieve multi-expert fusion. Thus, the post-edit representation $\hat{h}^{l_e}$ is obtained as follows:
\begin{gather}
\hat{h}^{l_e} = h^{l_e} + \!\!\!\!\!
\sum\limits_{(U_e,V_e,\hat{\psi}_{p_e})\in\hat{\mathcal{E}}}
\!\!\!\!\!\!\!\!
f_{sr}(\bar{\psi}_{p_i}, \hat{\psi}_{p_e}, \hat{\mathcal{E}}) \rho(h^{l_e} U_e^T)V_e \label{eq_reps_modify}\\
\text{s.t.} \quad f_{sr}(\bar{\psi}, \hat{\psi},\hat{\mathcal{E}}) = 
\sigma\left(\bar{\psi} \hat{\psi}^T\right)\frac{\exp(\bar{\psi} \hat{\psi}^T)}{\sum\limits_{(\_,\_,\hat{\psi}_{p_e})\in\hat{\mathcal{E}}}\exp(\bar{\psi} \hat{\psi}_{p_e}^T)} 
\end{gather}
where $\rho(\cdot)$ and $\sigma(\cdot)$ are ReLU and sigmoid, respectively. The inner products are rescaled by $\sqrt{d_m}$, which is omitted above. The $\hat{h}^{l_e}$ will proceed to complete the subsequent layer inference and generate the modified response. The Soft Routing function $f_{sr}(\cdot)$ multiplies absolute weights from the sigmoid and relative weights from the softmax. The absolute weights control the output strength of each expert based on similarity. The relative weights balance the similarity among the selected experts to constrain the scale of the fused residual output within 1, preventing the combined output from generating an excessively large norm.

\subsection{Training of LiveEdit}
The training primarily consists of two parts: the edit loss, which ensures that the generated MoEs effectively guide the VLLM to follow the editing instructions, and the routing loss, which ensures hard and soft MoE routing. Given a batch of edit samples $\mathcal{D}_e=\{(v_{e_b}, p_{e_b}, o_{e_b})\}_{b=1}^B$, and their corresponding sampled generality and locality samples $\mathcal{D}_g = \{(v_{g_b}, p_{g_b}, o_{g_b})\}_{b=1}^B$ and $\mathcal{D}_l = \{(v_{l_b}, p_{l_b}, o_{l_b})\}_{b=1}^B$, the losses are formulated as follows.

\subsubsection{Edit Loss}
We mix the experts for the entire batch of edit samples to simulate the scenario during inference, where hard routing leads to multiple experts. First, we obtain the $l_e$-th layer outputs $\{(h_{v_{e_b}}^{l_e}, h_{p_{e_b}}^{l_e}, h_{o_{e_b}}^{l_e})\}_{b=1}^B$ for each part of an edit sample in $\mathcal{D}_e$. Then, through Eqs.~\ref{eq_expert_generation} and \ref{eq_feature_extraction}, their corresponding experts and routing features can be obtained as $\{(U_{e_b}, V_{e_b})\}_{b=1}^B$ and $\{(\hat{\phi}_{v_{e_b}}, \hat{\psi}_{p_{e_b}})\}_{b=1}^B$. We define the expert set for soft routing fusion as $\hat{\mathcal{E}}=\{(U_{e_b}, V_{e_b}, \hat{\psi}_{p_{e_b}})\}_{b=1}^B$. Thus, for any input sample, its representation at layer $l_e$ will be modified as in Eq.~\ref{eq_reps_modify}. Defining the VLLM modified in this way as $f_{\theta_{\hat{\mathcal{E}}}}$, the edit loss is defined as follows:
\begin{gather}
    \ell_{\text{edit}} = \frac{1}{B} \sum\limits_{b=1}^B \left( \ell^{(b)}_{\text{rel}} + \ell^{(b)}_{\text{gen}} + \ell_{\text{loc}}^{(b)} \right)
\end{gather}
where
\begin{gather}
    \ell^{(b)}_{\text{rel}} = -\log f_{\theta_{\hat{\mathcal{E}}}}\left(o_{e_b} \mid v_{e_b},p_{e_b}\right)\\
    \ell^{(b)}_{\text{gen}} = -\log f_{\theta_{\hat{\mathcal{E}}}}\left(o_{g_b} \mid v_{g_b}, p_{g_b}\right)\\
    \ell^{(b)}_{\text{loc}} = 
    \text{KL}\left(
            f_{\theta}\left(o_{l_b} \mid v_{l_b}, p_{l_b}\right) \,||\,
            f_{\theta_{\hat{\mathcal{E}}}}\left(o_{l_b} \mid v_{l_b}, p_{l_b}\right)
        \right)
\end{gather}
Here, $\text{KL}$ denotes the Kullback-Leibler divergence.

\subsubsection{Routing Loss}
In the routing part, we maximize the feature similarity between samples within the generality domain, while minimizing the feature similarity between unrelated samples.
First, we randomly assign samples within the same generalization domain (i.e., edit samples and their corresponding generality samples) into two new sets, defined as
\(\hat{\mathcal{D}}_g = \{[\mathcal{D}_e^{(b)},\mathcal{D}_g^{(b)}]_{\pi_1^{(b)}}\}_{b=1}^B\), and
\(\bar{\mathcal{D}}_g = \{[\mathcal{D}_e^{(b)},\mathcal{D}_g^{(b)}]_{\pi_2^{(b)}}\}_{b=1}^B\).
Here, \(\pi_1^{(b)}, \pi_2^{(b)} \in [0,1]^B\) are the random integer vectors applied across the batch.
This approach equalizes the reliability and generality of samples in feature matching with the edited and input samples, enhancing routing robustness.
We use \(\hat{f}_{fe}\) to extract the routing features \(\{(\hat{\phi}_{g_b}, \hat{\psi}_{g_b})\}_{b=1}^B\) from \(\hat{\mathcal{D}}_g\) corresponding to edit end, 
and use \(\bar{f}_{fe}\) to extract the routing features \(\{(\bar{\phi}_{g_b}, \bar{\psi}_{g_b})\}_{b=1}^B\) and \(\{(\bar{\phi}_{l_b}, \bar{\psi}_{l_b})\}_{b=1}^B\) from \(\bar{\mathcal{D}}_g\) and \(\mathcal{D}_l\) corresponding to input end, respectively. 
The routing loss is formulated as follows:
\begin{gather}
\ell_{\text{route}} = \sum\limits_{b=1}^B\ell_{hr}^{(b)}+\ell_{sr_1}^{(b)} + \ell_{sr_2}^{(b)}
\end{gather}
\textbf{Hard Routing} loss \(\ell_{hr}^{(b)}\) is defined as follows:
\begin{equation}
\begin{aligned}
\ell_{hr}^{(b)} = f_{nce}(\bar{\phi}_{g_b}, \hat{\phi}_{g_b}, \hat{\Phi}\cup \{\bar{\phi}_{\Theta_{g_b}}\}) + \\
\;\;\;f_{nce}(\bar{\phi}_{l_b}, \bar{\phi}_{\Theta_{l_b}}, \hat{\Phi}\cup \{\bar{\phi}_{\Theta_{l_b}}\})
\end{aligned}
\end{equation}
where \(\hat{\Phi}=\{\hat{\phi}_{g_b}\}_{b=1}^B\). \(\bar{\phi}_{\Theta_{g_b}}\) and \(\bar{\phi}_{\Theta_{l_b}}\) represent the features extracted by the corresponding generality and locality data at the input end (as defined in Eq.\ref{eq_soft_routing_and_visual_sentinel}) from the vision sentinel. 
\(f_{nce}\) is the InfoNCE loss \cite{DBLP:journals/corr/abs-1807-03748} formulated as:
\begin{equation}
    f_{nce}(\alpha, \beta_+,\{\beta_j\}_{j=1}^n) = -\log \frac{\exp(\alpha \beta_+^T)}{\sum_{j=1}^n \exp(\alpha \beta_j^T)}
\end{equation}
We set the temperature to 1, which is omitted here.
The above loss function brings the key visual features of data within the generalization domain closer—even closer than those extracted using the visual sentinel. Meanwhile, it pushes the locality inputs further from the generalization domain, making them relatively closer to the features extracted from the visual sentinel.

\noindent\textbf{Soft Routing} includes absolute loss $\ell^{(b)}_{sr_1}$ and relative loss $\ell^{(b)}_{sr_2}$, defined as follows:
\begin{gather}
\ell_{sr_1}^{(b)} = -\log\sigma(\bar{\psi}_{g_b}\hat{\psi}_{g_b}^T) - \log(1-\sigma(\bar{\psi}_{g_b}\hat{\psi}_{\backslash g_{b}}^T)), \\ 
\ell_{sr_2}^{(b)} = 
f_{nce}\left(\bar{\psi}_{g_b}, \hat{\psi}_{g_b}, 
\{\hat{\psi}_{g_j}\}_{j=1}^B\cup
\{\hat{\psi}_{l_j}\}_{j=1}^B\right),
\end{gather}
where $\hat{\psi}_{\backslash g_b}$ represents a feature randomly selected from $\{\hat{\psi}_{g_j}\}_{j=1}^B\cup
\{\hat{\psi}_{l_j}\}_{j=1}^B \backslash \{\hat{\psi}_{g_{b}}\}$.
Thus, the total training loss is:
$\ell_{total} = \ell_{edit} + \ell_{route}$.
During training, the parameters of the VLLM, $f_{\theta}$, are frozen. The trainable modules are an experts generation module, $f_{eg}$, and two feature extraction modules, $\hat{f}_{fe}$ and $\bar{f}_{fe}$.

\begin{table*}[!tb]
    \scriptsize
    \centering
    \setlength{\tabcolsep}{3.83pt}
    \renewcommand{\arraystretch}{0.85}

\begin{tabular}{ccccccccccccccc}
\toprule
\multirow{2}{*}{\textbf{Baseline}}   & \multirow{2}{*}{\textbf{\# Edit}} & \multirow{2}{*}{\textbf{Editors}} & \multicolumn{6}{c}{\textbf{E-VQA}} & \multicolumn{6}{c}{\textbf{VLKEB}} \\
 &  &  & \textbf{Rel.}  & \textbf{T-Gen.} & \textbf{M-Gen.} & \textbf{T-Loc.} & \textbf{M-Loc.} & \textbf{Average}  & \textbf{Rel.}  & \textbf{T-Gen.} & \textbf{M-Gen.} & \textbf{T-Loc.} & \textbf{M-Loc.} & \textbf{Average}  \\
\midrule
\multirow{30}{*}{\begin{tabular}[c]{@{}c@{}}LLaVA-V1.5\\   (7B)\end{tabular}}  & \multirow{9}{*}{1} & FT-L   & 93.88 & 87.98  & 80.25  & 99.61  & 94.78  & 91.30 $_{(\pm0.42)}$ & 94.29 & 87.00  & 92.22  & 91.16  & 91.37  & 91.21 $_{(\pm1.09)}$ \\
 &  & FT-M   & 87.29 & 76.11  & 53.23  & 100.00 & 96.95  & 82.72 $_{(\pm1.05)}$ & 76.31 & 65.57  & 59.43  & 100.00 & 92.35  & 78.73 $_{(\pm0.76)}$ \\
 &  & MEND   & 91.23 & 90.05  & 91.29  & 91.02  & 90.22  & 90.76 $_{(\pm0.64)}$ & 92.13 & 91.28  & 90.22  & 89.19  & 90.13  & 90.59 $_{(\pm1.24)}$ \\
 &  & SERAC  & 89.33 & 83.72  & 84.97  & 82.05  & 23.78  & 72.77 $_{(\pm0.36)}$ & 89.77 & 89.11  & 87.92  & 66.68  & 14.20  & 69.54 $_{(\pm0.83)}$ \\
 &  & TP  & 35.95 & 36.12  & 28.65  & 93.87  & 97.61  & 58.44 $_{(\pm0.33)}$ & 50.77 & 55.70  & 51.65  & 87.93  & 90.43  & 67.30 $_{(\pm0.29)}$ \\
 &  & LTE & 94.16 & 93.57  & \textbf{93.59}  & 94.08  & 86.26  & 92.33 $_{(\pm1.56)}$ & 94.42 & 93.57  & 93.22  & 86.84  & 79.69  & 89.55 $_{(\pm1.41)}$ \\
 &  & RECIPE & 91.37 & 86.51  & 87.73  & 94.27  & 88.88  & 89.75 $_{(\pm1.13)}$ & 92.67 & 92.35  & 91.01  & 89.67  & 82.85  & 89.71 $_{(\pm0.57)}$ \\
 &  & LEMoE  & 93.60 & 92.77  & 89.99  & 99.28  & 96.98  & 94.52 $_{(\pm1.09)}$ & 94.85 & 93.09  & 91.67  & 87.03  & 87.88  & 90.90 $_{(\pm0.29)}$ \\
 &  & LiveEdit  & \textbf{94.28} & \textbf{94.51}  & 88.01  & \textbf{100.00} & \textbf{100.00} & \textbf{95.36} $_{(\pm0.57)}$ & \textbf{96.43} & \textbf{95.22}  & \textbf{93.72}  & \textbf{100.00} & \textbf{100.00} & \textbf{97.08} $_{(\pm0.62)}$ \\
\cmidrule{2-15}
 & \multirow{9}{*}{10}   & FT-L   & 90.57 & 84.14  & 73.21  & 95.56  & 81.50  & 85.00 $_{(\pm1.07)}$ & 88.05 & 85.32  & 85.23  & 74.53  & 85.74  & 83.77 $_{(\pm1.22)}$ \\
 &  & FT-M   & 84.90 & 73.53  & 49.99  & 100.00 & 55.98  & 72.88 $_{(\pm0.63)}$ & 68.63 & 57.57  & 56.56  & 100.00 & 82.99  & 73.15 $_{(\pm0.23)}$ \\
 &  & MEND   & 3.58  & 3.55   & 3.53   & 2.10   & 1.26   & 2.80  $_{(\pm0.02)}$ & 0.18  & 0.24   & 0.05   & 0.03   & 0.19   & 0.14  $_{(\pm0.00)}$ \\
 &  & SERAC  & 88.09 & 83.40  & 83.57  & 64.91  & 15.50  & 67.10 $_{(\pm0.92)}$ & 81.55 & 74.49  & 80.24  & 54.71  & 13.15  & 60.83 $_{(\pm0.98)}$ \\
 &  & TP  & 32.71 & 31.23  & 28.58  & 75.10  & 91.17  & 51.76 $_{(\pm0.60)}$ & 44.56 & 47.52  & 45.36  & 52.21  & 66.61  & 51.25 $_{(\pm0.69)}$ \\
 &  & LTE & 92.83 & 91.41  & \textbf{90.82}  & 86.38  & 85.52  & 89.39 $_{(\pm0.34)}$ & 90.06 & 81.52  & 88.11  & 83.40  & 81.48  & 84.91 $_{(\pm0.78)}$ \\
 &  & RECIPE & 90.22 & 85.92  & 86.24  & 90.34  & 88.11  & 88.17 $_{(\pm1.48)}$ & 83.92 & 76.23  & 82.84  & 86.33  & 83.69  & 82.60 $_{(\pm0.72)}$ \\
 &  & LEMoE  & 91.95 & 86.54  & 79.82  & 85.19  & 49.81  & 78.66 $_{(\pm1.03)}$ & 91.55 & 84.58  & 81.03  & 67.19  & 72.81  & 79.43 $_{(\pm0.52)}$ \\
 &  & LiveEdit  & \textbf{93.79} & \textbf{93.21}  & 86.42  & \textbf{100.00} & \textbf{100.00} & \textbf{94.68} $_{(\pm1.03)}$ & \textbf{95.54} & \textbf{94.52}  & \textbf{91.25}  & \textbf{100.00} & \textbf{100.00} & \textbf{96.26} $_{(\pm0.33)}$ \\
\cmidrule{2-15}
 & \multirow{9}{*}{1000} & FT-L   & 71.39 & 59.83  & 57.41  & 55.55  & 48.99  & 58.63 $_{(\pm0.17)}$ & 68.14 & 66.38  & 66.98  & 65.61  & 75.35  & 68.49 $_{(\pm0.32)}$ \\
 &  & FT-M   & 69.57 & 56.34  & 44.07  & 100.00 & 41.47  & 62.29 $_{(\pm0.40)}$ & 53.41 & 48.80  & 43.16  & 100.00 & 57.03  & 60.48 $_{(\pm0.50)}$ \\
 &  & MEND   & 0.04  & 0.05   & 0.05   & 0.08   & 0.09   & 0.06  $_{(\pm0.00)}$ & 0.03  & 0.05   & 0.07   & 0.06   & 0.08   & 0.06  $_{(\pm0.00)}$ \\
 &  & SERAC  & 85.57 & 75.58  & 82.01  & 62.46  & 15.69  & 64.26 $_{(\pm0.37)}$ & 60.93 & 56.49  & 60.06  & 52.94  & 15.04  & 49.09 $_{(\pm0.36)}$ \\
 &  & TP  & 16.56 & 16.80  & 15.65  & 7.28   & 15.60  & 14.38 $_{(\pm0.14)}$ & 5.46  & 4.81   & 5.51   & 2.77   & 7.19   & 5.15  $_{(\pm0.07)}$ \\
 &  & LTE & 83.93 & 82.55  & 81.34  & 83.97  & 73.09  & 80.98 $_{(\pm1.36)}$ & 64.51 & 56.26  & 64.80  & 80.85  & 76.52  & 68.59 $_{(\pm0.60)}$ \\
 &  & RECIPE & 87.00 & 76.81  & 83.09  & 86.95  & 87.03  & 84.18 $_{(\pm0.80)}$ & 62.00 & 56.84  & 61.50  & 85.37  & 82.07  & 69.56 $_{(\pm0.31)}$ \\
 &  & LEMoE  & 30.80 & 25.75  & 24.32  & 71.45  & 46.23  & 39.71 $_{(\pm0.23)}$ & 67.97 & 61.07  & 58.16  & 48.48  & 44.06  & 55.95 $_{(\pm0.36)}$ \\
 &  & LiveEdit  & \textbf{92.93} & \textbf{90.16}  & \textbf{84.30}  & \textbf{100.00} & \textbf{96.43}  & \textbf{92.76} $_{(\pm0.20)}$ & \textbf{92.22} & \textbf{83.97}  & \textbf{82.75}  & \textbf{100.00} & \textbf{100.00} & \textbf{91.79} $_{(\pm0.55)}$ \\
\midrule
\multirow{20}{*}{\begin{tabular}[c]{@{}c@{}}BLIP2-OPT\\   (2.7B)\end{tabular}} & \multirow{9}{*}{1} & FT-L   & 52.86 & 48.80  & 32.94  & 98.24  & 94.27  & 65.42 $_{(\pm0.69)}$ & 54.31 & 54.27  & 54.08  & 98.40  & 94.37  & 71.09 $_{(\pm1.05)}$ \\
 &  & FT-M   & 91.70 & 87.24  & 33.30  & 100.00 & 85.22  & 79.49 $_{(\pm0.72)}$ & 92.64 & 80.97  & 63.62  & 100.00 & 83.02  & 84.05 $_{(\pm0.70)}$ \\
 &  & MEND   & 93.13 & 92.76  & 93.07  & 92.00  & 75.81  & 89.35 $_{(\pm0.93)}$ & 94.91 & 93.81  & 93.84  & 94.98  & 86.54  & 92.82 $_{(\pm0.82)}$ \\
 &  & SERAC  & 88.39 & 84.50  & 84.25  & 85.82  & 26.00  & 73.79 $_{(\pm1.01)}$ & 87.95 & 84.67  & 85.20  & 68.10  & 17.75  & 68.73 $_{(\pm0.97)}$ \\
 &  & TP  & 70.14 & 65.80  & 53.05  & 98.11  & 85.33  & 74.49 $_{(\pm0.38)}$ & 50.98 & 49.47  & 50.88  & 94.76  & 78.57  & 64.93 $_{(\pm1.02)}$ \\
 &  & LTE & 95.74 & 93.86  & 86.90  & 97.93  & 87.97  & 92.48 $_{(\pm0.70)}$ & 94.13 & 91.93  & 92.23  & 93.89  & 92.27  & 92.89 $_{(\pm1.01)}$ \\
 &  & RECIPE & 89.42 & 86.24  & 87.53  & 99.87  & 89.16  & 90.45 $_{(\pm1.46)}$ & 92.38 & 89.74  & 89.17  & 97.13  & 94.46  & 92.58 $_{(\pm1.16)}$ \\
 &  & LEMoE  & 93.56 & 92.23  & 91.40  & 98.50  & 85.21  & 92.18 $_{(\pm0.73)}$ & 94.59 & 93.14  & 92.37  & 94.53  & 61.53  & 87.23 $_{(\pm0.34)}$ \\
 &  & LiveEdit  & \textbf{96.67} & \textbf{94.20}  & \textbf{93.82}  & \textbf{100.00} & \textbf{100.00} & \textbf{96.94} $_{(\pm1.32)}$ & \textbf{98.77} & \textbf{98.08}  & \textbf{94.89}  & \textbf{100.00} & \textbf{100.00} & \textbf{98.35} $_{(\pm1.58)}$ \\
\cmidrule{2-15}
 & \multirow{9}{*}{1000} & FT-L   & 45.10 & 34.62  & 35.42  & 48.42  & 41.24  & 40.96 $_{(\pm0.29)}$ & 55.39 & 54.34  & 53.87  & 50.80  & 54.00  & 53.68 $_{(\pm0.80)}$ \\
 &  & FT-M   & 40.40 & 31.46  & 27.85  & 100.00 & 27.44  & 45.43 $_{(\pm0.68)}$ & 47.03 & 49.68  & 46.99  & 100.00 & 41.41  & 57.02 $_{(\pm0.13)}$ \\
 &  & MEND   & 15.84 & 14.35  & 17.73  & 91.74  & 70.17  & 41.97 $_{(\pm0.12)}$ & 37.22 & 38.03  & 37.19  & 91.49  & 84.10  & 57.61 $_{(\pm0.58)}$ \\
 &  & SERAC  & 83.35 & 70.80  & 80.32  & 67.66  & 13.13  & 63.05 $_{(\pm0.87)}$ & 53.58 & 45.78  & 52.42  & 56.81  & 16.90  & 45.10 $_{(\pm0.38)}$ \\
 &  & TP  & 20.63 & 15.09  & 18.41  & 8.65   & 8.25   & 14.21 $_{(\pm0.18)}$ & 24.36 & 24.21  & 24.25  & 16.37  & 19.96  & 21.83 $_{(\pm0.14)}$ \\
 &  & LTE & 89.32 & 82.82  & 81.51  & 94.86  & 69.83  & 83.67 $_{(\pm1.05)}$ & 61.67 & 51.05  & 61.60  & 94.78  & 90.94  & 72.01 $_{(\pm0.66)}$ \\
 &  & RECIPE & 84.99 & 74.20  & 82.04  & 96.82  & 87.73  & 85.16 $_{(\pm1.32)}$ & 54.64 & 46.54  & 54.10  & 94.60  & 96.93  & 69.37 $_{(\pm1.04)}$ \\
 &  & LEMoE  & 19.73 & 17.34  & 18.22  & 72.01  & 31.06  & 31.67 $_{(\pm0.14)}$ & 34.74 & 33.43  & 32.05  & 55.55  & 50.04  & 41.16 $_{(\pm0.58)}$ \\
 &  & LiveEdit  & \textbf{94.42} & \textbf{91.98}  & \textbf{84.65}  & \textbf{100.00} & \textbf{97.38}  & \textbf{93.69} $_{(\pm0.67)}$ & \textbf{97.00} & \textbf{91.92}  & \textbf{87.53}  & \textbf{100.00} & \textbf{100.00} & \textbf{95.29} $_{(\pm1.48)}$ \\
\bottomrule
\end{tabular}

    \caption{Partial results of lifelong edit performance for BLIP2-OPT and LLaVA-V1.5 on the E-VQA and VLKEB datasets. Due to space limitations, please refer to Appendix \ref{appendix_complete_main_exps} for the complete results, including those for the E-IC dataset and the MiniGPT-4 model. ``Rel.'', ``T/M-Gen.'' and ``T/M-Loc.'' stand for reliability, text/modal generality, and text/modal locality, respectively. ``\# Edit'' indicates the number of edits. The t-tests demonstrate our improvements are statistically significant with $p < 0.05$ level.}
    \label{tab_main_exp}
    \vspace{-1.8em}
\end{table*}

\section{Experiments}
\subsection{Experimental Settings}

\noindent\textbf{Datasets:} Following \cite{MMEdit}, we use E-VQA (Editing Visual Question Answering) and E-IC (Editing Image Caption) as evaluation datasets. Additionally, we incorporate VLKEB \cite{VLKEB}, which is composed of real images to better represent real-world scenarios.

\noindent\textbf{VLLM Backbones:} For comprehensive evaluation, we select VLLM backbones based on both model architecture and parameter scale, including BLIP2-OPT (2.7B) \cite{BLIP2}, LLaVA-V1.5 (7B) \cite{llava}, and MiniGPT-4 (7B) \cite{MiniGPT-4}.

\noindent\textbf{Baseline Editors:} To our knowledge, there are currently no editors specifically designed for lifelong VLLM editing. Therefore, following \cite{MMEdit}, in addition to the basic FT-L and FT-M, which respectively fine-tune the final layer of the LLM and the visual encoder, we adapt LLM-based editing techniques to VLLM. These include MEND \cite{MEND}, TP \cite{T-Patcher}, LTE \cite{LTE}, RECIPE \cite{RECIPE}, and LEMoE \cite{LEMOE}.

For details on the experimental setup, model hyper-parameters, and training specifics, please refer to Appendix \ref{Appendix_detail_exp_settings}. Building on the experimental settings above, we conduct a comprehensive evaluation of edit performance and perform an in-depth analysis of LiveEdit's internals.

\subsection{General Performance of Lifelong Editing}

Table \ref{tab_main_exp} shows partial results of lifelong editing experiments.
In single-edit scenarios, our method generally achieves optimal performance. 
Methods such as FT-L, FT-M, and MEND \cite{MEND} try to modify original model parameters and perform well initially. However, their performance deteriorates with more edits due to overfitting and cumulative parameter scaling \cite{WILKE, DBLP:journals/corr/abs-2408-07413}. 
TP \cite{T-Patcher} addresses parameter scaling by introducing additional neurons, but a single neuron can't encapsulate each edit's visual information. Retrieval-based methods (SERAC \cite{SERAC}, LTE \cite{LTE}, RECIPE \cite{RECIPE}) remain robust in lifelong editing by decoupling edits from model parameters, yet struggle with semantically similar but visually different edit samples.
LEMoE \cite{LEMOE}, a MoE-based editor, excels with few edits but suffers from issues like disruptive greedy routing and limited generalization due to overfitting experts to batch edits. Our method, LiveEdit, surpasses in edit performance. Contrastive learning-based expert routing resolves LEMoE's issues, enhancing both edit speed and generalization. 
By decoupling edit samples into independent experts and applying a fusion strategy for vision-related experts, we prevent the semantic conflicts LEMoE encounters from its sequential batching. Importantly, our method maintains nearly 100\% locality performance even with increasing edits.
Three main factors contribute to our performance: First, hard routing filters out visually irrelevant experts. Second, soft routing scales influence by assigning lower scales to textually irrelevant experts. Finally, the locality edit loss further confines experts' influence on response adaptation.

\begin{figure}[!t]
    \centering
    \includegraphics[width=1.\columnwidth]{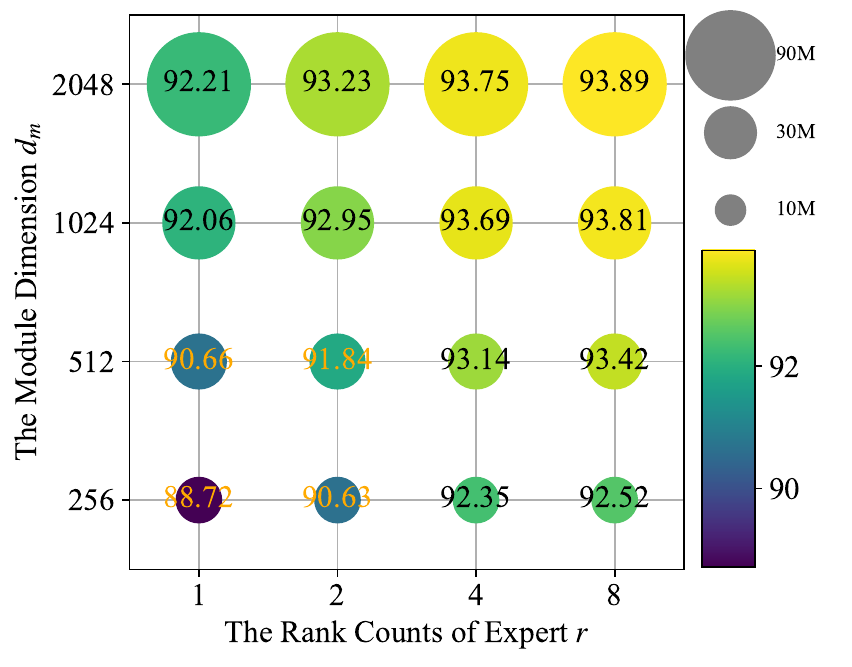}
    \vspace{-2.0em}
\caption{
The impact of module dimension $d_m$ and expert rank $r$ on LiveEdit's edit performance. Experiments are conducted on the E-VQA dataset, with 1,000 edits on BLIP2. Circle size represents LiveEdit's training parameters, and color intensity indicates the average edit performance across five metrics.
} 
    \label{fig_module_dim_trade_off_rank}
    \vspace{-1.5em}
\end{figure}

\subsection{Hyper-Parameter Search}
We conducted a comprehensive hyper-parameters search for LiveEdit to select the most suitable combination. The most important hyper-parameters are discussed below.

\begin{figure}[!t]
    \centering
    \includegraphics[width=1.\columnwidth]{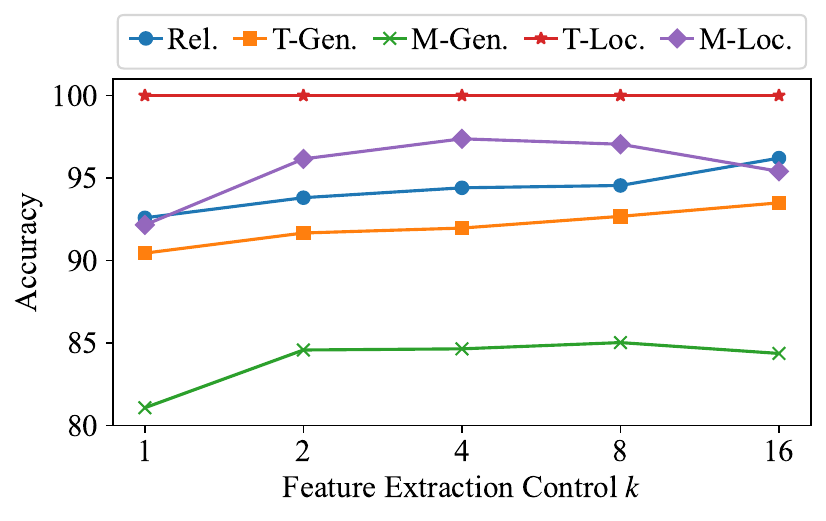}
    \vspace{-2.5em}
\caption{The dimension control parameter $k$ for feature extraction.} 
    \label{fig_Feature_Extraction_Control}
    \vspace{-1.0em}
\end{figure}

\noindent\textbf{Trade-off Between Model Scale and Edit Performance}: Figure \ref{fig_module_dim_trade_off_rank} reports the effects of different combinations of module dimension $d_m$ and expert rank $r$ on edit performance and model scale. The $d_m$ has a significant impact on the parameter count of LiveEdit. In terms of edit performance, both $d_m$ and $r$ have substantial effects. 
Increasing one while keeping the other fixed improves edit performance, though the improvement gradually becomes flat.
Additionally, combinations along the diagonal show generally consistent performance. 
Regarding model scale, $d_m$ predominantly influences variation, while $r$ has minimal impact. 
Based on this analysis, the optimal configuration strategy is to select an appropriate $d_m$ and maximize $r$ as much as possible. 
However, since $r$ linearly controls the growth rate of the expert repository, specific choices should also consider the memory needed to store the expert repository.
Additionally, we expand the dimension control parameter $k$ for feature extraction, as shown in Figure \ref{fig_Feature_Extraction_Control}. Increasing $k$ enhances the feature extraction capability, but an excessively high $k$ may introduce noise, leading to incorrect matches, such as reduced modal locality.

\begin{figure}[t]
    \centering
    \includegraphics[width=1.\columnwidth]{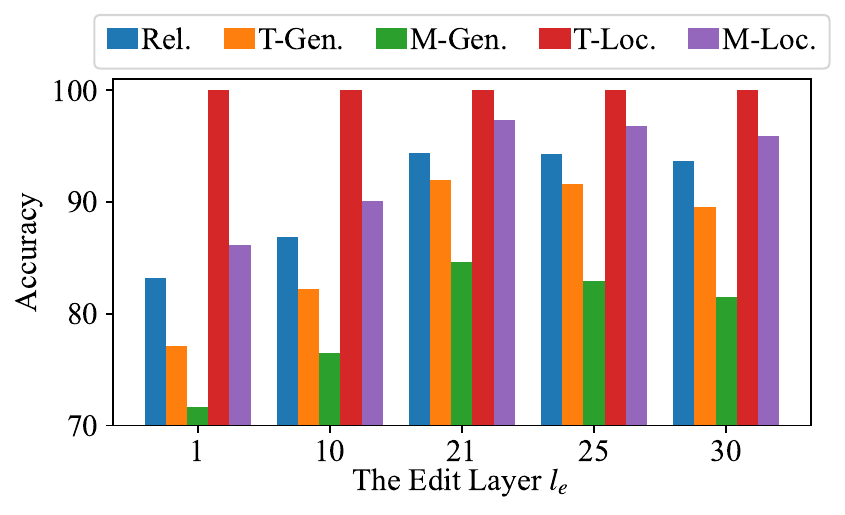}
    \vspace{-2.5em}
    \caption{
    Impact of LiveEdit attached layer index $l_e$. 
    Results of 1,000 edits for BLIP2 on E-VQA dataset are reported.
    } 
    \label{fig_module_put_layer}
    \vspace{-0.5em}
\end{figure}

\noindent\textbf{The Attached Layer of LiveEdit}: 
Figure \ref{fig_module_put_layer} shows the impact of the layer attached by LiveEdit on edit performance. It can be seen that as the layer depth increases, edit performance improves, reaching a peak at 21 layers. After this point, edit performance slightly declines. 
We speculate that transformers typically perform semantic understanding in the early layers and response generation in the later layers \cite{DBLP:conf/acl/JawaharSS19, ROME, DBLP:journals/corr/abs-2408-09916}. 
Attaching to an earlier layer prevents LiveEdit from leveraging VLLM's semantic understanding capabilities to enhance the feature extraction process.
For too deep layers, VLLM has largely stabilized the predictive tendencies of the response, making it more challenging for LiveEdit to adapt. The above results also align with the attribution conclusions of \cite{DBLP:journals/corr/abs-2408-09916}.

\begin{table}[!tb]
    \footnotesize
    \centering
    \setlength{\tabcolsep}{4.5pt}
    \renewcommand{\arraystretch}{0.85}

\begin{tabular}{lcccccc}
\toprule
\textbf{Settings} & \textbf{Rel.}  & \textbf{T-Gen.} & \textbf{M-Gen.} & \textbf{T-Loc.} & \textbf{M-Loc.} & \textbf{Average} \\
\midrule
N/A               & 20.57          & 19.03           & 14.17           & 100.00          & 100.00          & 50.75            \\
\midrule
LiveEdit          & \textbf{94.42} & \textbf{91.98}  & \textbf{84.65}  & \textbf{100.00} & \textbf{97.38}  & \textbf{93.69}   \\
- $\ell_{sr1}$    & 87.93          & 83.77           & 73.40           & 100.00          & 76.93           & 84.41            \\
- $\ell_{sr2}$    & 89.21          & 85.46           & 77.34           & 100.00          & 91.69           & 88.74            \\
- SR              & 88.92          & 81.49           & 70.94           & 100.00          & 75.66           & 83.40            \\
HR*               & 93.60          & 88.50           & 80.37           & 100.00          & 84.77           & 89.45           \\
\bottomrule
\end{tabular}

    \caption{Ablation study of LiveEdit.
    }
    \label{tab_ablation_study}
    \vspace{-1.5em}
\end{table}

\subsection{Ablation Study}
Table \ref{tab_ablation_study} presents the ablation results for LiveEdit, where 1000 edits are applied to BLIP2 on the E-VQA dataset. The removal of soft routing (- SR, which directly averages experts) results in a significant performance drop for LiveEdit. 
We can also observe that a large portion of this loss originates from the removal of the absolute soft routing loss $\ell_{sr_1}$, particularly impacting modal locality. 
This is because all hard-routed experts are assigned a weight to adapt the representation, even if they are irrelevant to the input prompt, ensuring that the sum of adaptation strengths equals 1. 
Similarly, removing the relative soft routing loss $\ell_{sr_2}$ also leads to performance degradation, as the combined absolute weights of multiple experts may exceed 1 during testing, resulting in excessively high values. 
HR* modifies the visual extraction that leverages prompt semantics by directly compressing the entire visual representation for hard routing. This introduces additional visual noise, leading to the selection of more irrelevant experts, which in turn reduces the efficacy of subsequent steps.

\begin{figure}[!t]
    \centering
    \includegraphics[width=1.\columnwidth]{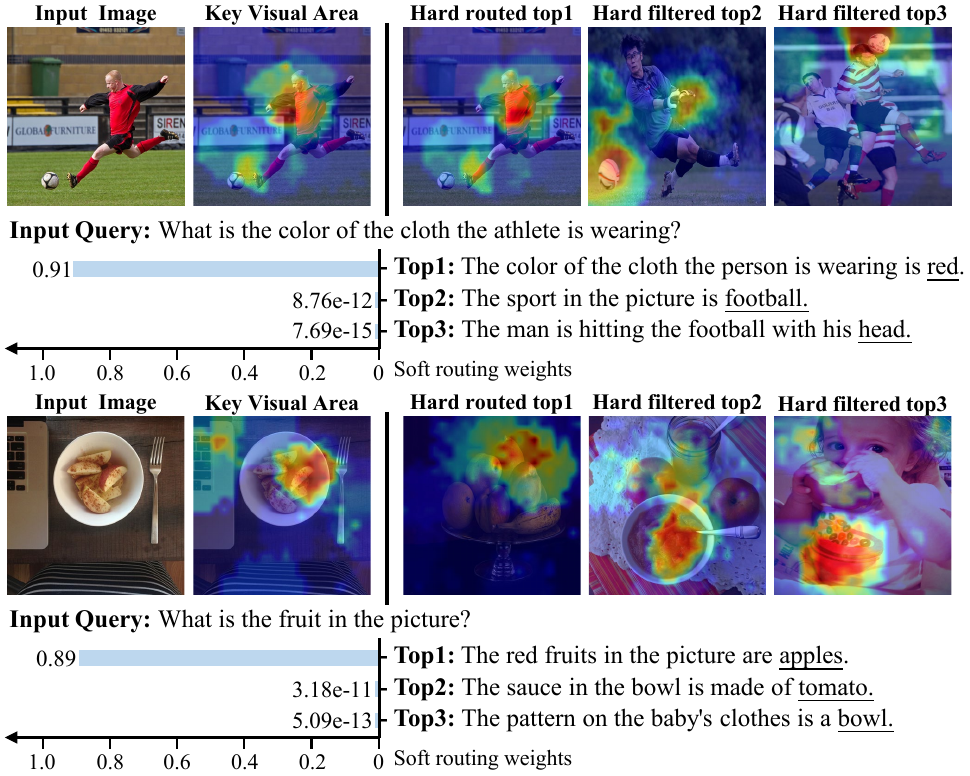}
    \caption{
    Instance analysis. The left side shows the model input to LLaVa after editing with LiveEdit. The right side displays the top 3 experts based on hard routing. The heatmaps represent the visual regions focused on when extracting key visual features. The bar chart below shows fusion weights from soft routing. 
    Please refer to Appendix \ref{appendix_hard_Instance_analysis} for more analysis.
    } 
    \label{fig_instance_analysis}
    \vspace{-1.0em}
\end{figure}

\subsection{Instance Analysis}
We conducted an instance analysis of LiveEdit, as shown in Figure \ref{fig_instance_analysis}. We perform 100 edits on LLaVa, including samples partially related in visual context to the incoming input samples. The figure reports the top 3 hard routing results for two inputs, as well as the fusion weights these experts received in soft routing. 
Using a perturbation-based attribution tool~\cite{DBLP:conf/iccv/FongV17, DBLP:journals/corr/abs-2408-09916}, we visualize the visual regions focused on for key visual feature extraction. 
It can be observed that the highlighted regions are closely aligned with the prompt semantics, which filters out irrelevant visual noise and benefits hard routing. The bar chart indicates that experts unrelated to the input query receive very low fusion weights, significantly benefiting the locality of edits.

\section{Conclusion}

In conclusion, we introduce LiveEdit to bridge the gap between LLM and VLLM editing, with an editing expert generator and a combination of hard/soft routers. The framework successfully addresses the limitation of exiting editors in the VLLM lifelong editing scenarios.
Our benchmark and extensive experiments confirm its superiority and highlight the effectiveness of each component, which support more accurate and adaptable VLLM editing in real-world applications.

\noindent\textbf{Acknowledgments.}
This work is supported by the National Key R\&D Program of China (2022ZD0120302).

{
    \small
    \bibliographystyle{ieeetr}
    \bibliography{main}

\begin{thebibliography}{10}

\bibitem{DBLP:journals/corr/abs-2302-13971}
H.~Touvron, T.~Lavril, G.~Izacard, X.~Martinet, M.~Lachaux, T.~Lacroix, B.~Rozi{\`{e}}re, N.~Goyal, E.~Hambro, F.~Azhar, A.~Rodriguez, A.~Joulin, E.~Grave, and G.~Lample, ``Llama: Open and efficient foundation language models,'' {\em CoRR}, vol.~abs/2302.13971, 2023.

\bibitem{DBLP:journals/fi/RoumeliotisT23}
K.~I. Roumeliotis and N.~D. Tselikas, ``Chatgpt and open-ai models: {A} preliminary review,'' {\em Future Internet}, vol.~15, no.~6, p.~192, 2023.

\bibitem{DBLP:conf/iclr/ZengLDWL0YXZXTM23}
A.~Zeng, X.~Liu, Z.~Du, Z.~Wang, H.~Lai, M.~Ding, Z.~Yang, Y.~Xu, W.~Zheng, X.~Xia, W.~L. Tam, Z.~Ma, Y.~Xue, J.~Zhai, W.~Chen, Z.~Liu, P.~Zhang, Y.~Dong, and J.~Tang, ``{GLM-130B:} an open bilingual pre-trained model,'' in {\em The Eleventh International Conference on Learning Representations, {ICLR} 2023, Kigali, Rwanda, May 1-5, 2023}, 2023.

\bibitem{BLIP2}
J.~Li, D.~Li, S.~Savarese, and S.~C.~H. Hoi, ``{BLIP-2:} bootstrapping language-image pre-training with frozen image encoders and large language models,'' in {\em International Conference on Machine Learning, {ICML} 2023, 23-29 July 2023, Honolulu, Hawaii, {USA}}, vol.~202 of {\em Proceedings of Machine Learning Research}, pp.~19730--19742, 2023.

\bibitem{DBLP:conf/nips/Dai0LTZW0FH23}
W.~Dai, J.~Li, D.~Li, A.~M.~H. Tiong, J.~Zhao, W.~Wang, B.~Li, P.~Fung, and S.~C.~H. Hoi, ``Instructblip: Towards general-purpose vision-language models with instruction tuning,'' in {\em Advances in Neural Information Processing Systems 36: Annual Conference on Neural Information Processing Systems 2023, NeurIPS 2023, New Orleans, LA, USA, December 10 - 16, 2023}, 2023.

\bibitem{DBLP:journals/corr/abs-2308-12966}
J.~Bai, S.~Bai, S.~Yang, S.~Wang, S.~Tan, P.~Wang, J.~Lin, C.~Zhou, and J.~Zhou, ``Qwen-vl: {A} frontier large vision-language model with versatile abilities,'' {\em CoRR}, vol.~abs/2308.12966, 2023.

\bibitem{DBLP:journals/corr/abs-2309-15112}
P.~Zhang, X.~Dong, B.~Wang, Y.~Cao, C.~Xu, L.~Ouyang, Z.~Zhao, S.~Ding, S.~Zhang, H.~Duan, W.~Zhang, H.~Yan, X.~Zhang, W.~Li, J.~Li, K.~Chen, C.~He, X.~Zhang, Y.~Qiao, D.~Lin, and J.~Wang, ``Internlm-xcomposer: {A} vision-language large model for advanced text-image comprehension and composition,'' {\em CoRR}, vol.~abs/2309.15112, 2023.

\bibitem{DBLP:conf/acl/JangYYCLLS23}
J.~Jang, D.~Yoon, S.~Yang, S.~Cha, M.~Lee, L.~Logeswaran, and M.~Seo, ``Knowledge unlearning for mitigating privacy risks in language models,'' in {\em Proceedings of the 61st Annual Meeting of the Association for Computational Linguistics (Volume 1: Long Papers), {ACL} 2023, Toronto, Canada, July 9-14, 2023}, pp.~14389--14408, 2023.

\bibitem{DBLP:journals/corr/abs-2309-11852}
Y.~Ishibashi and H.~Shimodaira, ``Knowledge sanitization of large language models,'' {\em CoRR}, vol.~abs/2309.11852, 2023.

\bibitem{DBLP:conf/iclr/LiLCZLW0024}
Y.~Li, T.~Li, K.~Chen, J.~Zhang, S.~Liu, W.~Wang, T.~Zhang, and Y.~Liu, ``Badedit: Backdooring large language models by model editing,'' in {\em The Twelfth International Conference on Learning Representations, {ICLR} 2024, Vienna, Austria, May 7-11, 2024}, 2024.

\bibitem{DBLP:conf/acl/Wang0XXDYZY0C24}
M.~Wang, N.~Zhang, Z.~Xu, Z.~Xi, S.~Deng, Y.~Yao, Q.~Zhang, L.~Yang, J.~Wang, and H.~Chen, ``Detoxifying large language models via knowledge editing,'' in {\em Proceedings of the 62nd Annual Meeting of the Association for Computational Linguistics (Volume 1: Long Papers), {ACL} 2024, Bangkok, Thailand, August 11-16, 2024}, pp.~3093--3118, 2024.

\bibitem{DBLP:conf/acl/YuJKYJ23}
C.~Yu, S.~Jeoung, A.~Kasi, P.~Yu, and H.~Ji, ``Unlearning bias in language models by partitioning gradients,'' in {\em Findings of the Association for Computational Linguistics: {ACL} 2023, Toronto, Canada, July 9-14, 2023}, pp.~6032--6048, 2023.

\bibitem{DBLP:conf/iclr/LimisiewiczMM24}
T.~Limisiewicz, D.~Marecek, and T.~Musil, ``Debiasing algorithm through model adaptation,'' in {\em The Twelfth International Conference on Learning Representations, {ICLR} 2024, Vienna, Austria, May 7-11, 2024}, 2024.

\bibitem{DBLP:journals/corr/abs-2408-11843}
R.~Chen, Y.~Li, J.~Yang, J.~T. Zhou, and Z.~Liu, ``Editable fairness: Fine-grained bias mitigation in language models,'' {\em CoRR}, vol.~abs/2408.11843, 2024.

\bibitem{DBLP:conf/iclr/ChuangXLKGH24}
Y.~Chuang, Y.~Xie, H.~Luo, Y.~Kim, J.~R. Glass, and P.~He, ``Dola: Decoding by contrasting layers improves factuality in large language models,'' in {\em The Twelfth International Conference on Learning Representations, {ICLR} 2024, Vienna, Austria, May 7-11, 2024}, 2024.

\bibitem{DBLP:conf/acl/ZhangY024}
S.~Zhang, T.~Yu, and Y.~Feng, ``Truthx: Alleviating hallucinations by editing large language models in truthful space,'' in {\em Proceedings of the 62nd Annual Meeting of the Association for Computational Linguistics (Volume 1: Long Papers), {ACL} 2024, Bangkok, Thailand, August 11-16, 2024}, pp.~8908--8949, 2024.

\bibitem{ROME}
K.~Meng, D.~Bau, A.~Andonian, and Y.~Belinkov, ``Locating and editing factual associations in {GPT},'' in {\em Advances in Neural Information Processing Systems 35: Annual Conference on Neural Information Processing Systems 2022, NeurIPS 2022, New Orleans, LA, USA, November 28 - December 9, 2022}, 2022.

\bibitem{MEMIT}
K.~Meng, A.~S. Sharma, A.~J. Andonian, Y.~Belinkov, and D.~Bau, ``Mass-editing memory in a transformer,'' in {\em The Eleventh International Conference on Learning Representations, {ICLR} 2023, Kigali, Rwanda, May 1-5, 2023}, 2023.

\bibitem{LTE}
Y.~Jiang, Y.~Wang, C.~Wu, W.~Zhong, X.~Zeng, J.~Gao, L.~Li, X.~Jiang, L.~Shang, R.~Tang, Q.~Liu, and W.~Wang, ``Learning to edit: Aligning llms with knowledge editing,'' in {\em Proceedings of the 62nd Annual Meeting of the Association for Computational Linguistics (Volume 1: Long Papers), {ACL} 2024, Bangkok, Thailand, August 11-16, 2024}, pp.~4689--4705, 2024.

\bibitem{RECIPE}
Q.~Chen, T.~Zhang, X.~He, D.~Li, C.~Wang, L.~Huang, and X.~Hui, ``Lifelong knowledge editing for {LLM}s with retrieval-augmented continuous prompt learning,'' in {\em Proceedings of the 2024 Conference on Empirical Methods in Natural Language Processing}, pp.~13565--13580, Nov. 2024.

\bibitem{WILKE}
C.~Hu, P.~Cao, Y.~Chen, K.~Liu, and J.~Zhao, ``Wilke: Wise-layer knowledge editor for lifelong knowledge editing,'' in {\em Findings of the Association for Computational Linguistics, {ACL} 2024, Bangkok, Thailand and virtual meeting, August 11-16, 2024}, pp.~3476--3503, 2024.

\bibitem{MMEdit}
S.~Cheng, B.~Tian, Q.~Liu, X.~Chen, Y.~Wang, H.~Chen, and N.~Zhang, ``Can we edit multimodal large language models?,'' in {\em Proceedings of the 2023 Conference on Empirical Methods in Natural Language Processing, {EMNLP} 2023, Singapore, December 6-10, 2023}, pp.~13877--13888, 2023.

\bibitem{DBLP:journals/corr/abs-2406-04236}
S.~Basu, M.~Grayson, C.~Morrison, B.~Nushi, S.~Feizi, and D.~Massiceti, ``Understanding information storage and transfer in multi-modal large language models,'' {\em CoRR}, vol.~abs/2406.04236, 2024.

\bibitem{DBLP:journals/corr/abs-2408-09916}
Q.~Chen, T.~Zhang, C.~Wang, X.~He, D.~Wang, and T.~Liu, ``Attribution analysis meets model editing: Advancing knowledge correction in vision language models with visedit,'' {\em CoRR}, vol.~abs/2408.09916, 2024.

\bibitem{GRACE}
T.~Hartvigsen, S.~Sankaranarayanan, H.~Palangi, Y.~Kim, and M.~Ghassemi, ``Aging with {GRACE:} lifelong model editing with discrete key-value adaptors,'' in {\em Advances in Neural Information Processing Systems 36: Annual Conference on Neural Information Processing Systems 2023, NeurIPS 2023, New Orleans, LA, USA, December 10 - 16, 2023}, 2023.

\bibitem{DBLP:journals/neco/JacobsJNH91}
R.~A. Jacobs, M.~I. Jordan, S.~J. Nowlan, and G.~E. Hinton, ``Adaptive mixtures of local experts,'' {\em Neural Comput.}, vol.~3, no.~1, pp.~79--87, 1991.

\bibitem{DBLP:conf/iclr/ShazeerMMDLHD17}
N.~Shazeer, A.~Mirhoseini, K.~Maziarz, A.~Davis, Q.~V. Le, G.~E. Hinton, and J.~Dean, ``Outrageously large neural networks: The sparsely-gated mixture-of-experts layer,'' in {\em 5th International Conference on Learning Representations, {ICLR} 2017, Toulon, France, April 24-26, 2017, Conference Track Proceedings}, 2017.

\bibitem{llava}
H.~Liu, C.~Li, Q.~Wu, and Y.~J. Lee, ``Visual instruction tuning,'' in {\em Advances in Neural Information Processing Systems 36: Annual Conference on Neural Information Processing Systems 2023, NeurIPS 2023, New Orleans, LA, USA, December 10 - 16, 2023}, 2023.

\bibitem{MiniGPT-4}
D.~Zhu, J.~Chen, X.~Shen, X.~Li, and M.~Elhoseiny, ``Minigpt-4: Enhancing vision-language understanding with advanced large language models,'' in {\em The Twelfth International Conference on Learning Representations, {ICLR} 2024, Vienna, Austria, May 7-11, 2024}, 2024.

\bibitem{VLKEB}
H.~Huang, H.~Zhong, T.~Yu, Q.~Liu, S.~Wu, L.~Wang, and T.~Tan, ``Vlkeb: A large vision-language model knowledge editing benchmark,'' 2024.

\bibitem{DBLP:journals/tist/ChangWWWYZCYWWYZCYYX24}
Y.~Chang, X.~Wang, J.~Wang, Y.~Wu, L.~Yang, K.~Zhu, H.~Chen, X.~Yi, C.~Wang, Y.~Wang, W.~Ye, Y.~Zhang, Y.~Chang, P.~S. Yu, Q.~Yang, and X.~Xie, ``A survey on evaluation of large language models,'' {\em {ACM} Trans. Intell. Syst. Technol.}, vol.~15, no.~3, pp.~39:1--39:45, 2024.

\bibitem{DBLP:conf/bigdataconf/WuGCWY23}
J.~Wu, W.~Gan, Z.~Chen, S.~Wan, and P.~S. Yu, ``Multimodal large language models: {A} survey,'' in {\em {IEEE} International Conference on Big Data, BigData 2023, Sorrento, Italy, December 15-18, 2023}, pp.~2247--2256, 2023.

\bibitem{DBLP:conf/iclr/DosovitskiyB0WZ21}
A.~Dosovitskiy, L.~Beyer, A.~Kolesnikov, D.~Weissenborn, X.~Zhai, T.~Unterthiner, M.~Dehghani, M.~Minderer, G.~Heigold, S.~Gelly, J.~Uszkoreit, and N.~Houlsby, ``An image is worth 16x16 words: Transformers for image recognition at scale,'' in {\em 9th International Conference on Learning Representations, {ICLR} 2021, Virtual Event, Austria, May 3-7, 2021}, 2021.

\bibitem{DBLP:journals/corr/abs-2306-16527}
H.~Lauren{\c{c}}on, L.~Saulnier, L.~Tronchon, S.~Bekman, A.~Singh, A.~Lozhkov, T.~Wang, S.~Karamcheti, A.~M. Rush, D.~Kiela, M.~Cord, and V.~Sanh, ``{OBELISC:} an open web-scale filtered dataset of interleaved image-text documents,'' {\em CoRR}, vol.~abs/2306.16527, 2023.

\bibitem{DBLP:conf/nips/AlayracDLMBHLMM22}
J.~Alayrac, J.~Donahue, P.~Luc, A.~Miech, I.~Barr, Y.~Hasson, K.~Lenc, A.~Mensch, K.~Millican, M.~Reynolds, R.~Ring, E.~Rutherford, S.~Cabi, T.~Han, Z.~Gong, S.~Samangooei, M.~Monteiro, J.~L. Menick, S.~Borgeaud, A.~Brock, A.~Nematzadeh, S.~Sharifzadeh, M.~Binkowski, R.~Barreira, O.~Vinyals, A.~Zisserman, and K.~Simonyan, ``Flamingo: a visual language model for few-shot learning,'' in {\em Advances in Neural Information Processing Systems 35: Annual Conference on Neural Information Processing Systems 2022, NeurIPS 2022, New Orleans, LA, USA, November 28 - December 9, 2022}, 2022.

\bibitem{DBLP:conf/cvpr/GoyalKSBP17}
Y.~Goyal, T.~Khot, D.~Summers{-}Stay, D.~Batra, and D.~Parikh, ``Making the {V} in {VQA} matter: Elevating the role of image understanding in visual question answering,'' in {\em 2017 {IEEE} Conference on Computer Vision and Pattern Recognition, {CVPR} 2017, Honolulu, HI, USA, July 21-26, 2017}, pp.~6325--6334, 2017.

\bibitem{DBLP:conf/cvpr/HudsonM19}
D.~A. Hudson and C.~D. Manning, ``{GQA:} {A} new dataset for real-world visual reasoning and compositional question answering,'' in {\em {IEEE} Conference on Computer Vision and Pattern Recognition, {CVPR} 2019, Long Beach, CA, USA, June 16-20, 2019}, pp.~6700--6709, 2019.

\bibitem{chiang2023vicuna}
W.-L. Chiang, Z.~Li, Z.~Lin, Y.~Sheng, Z.~Wu, H.~Zhang, L.~Zheng, S.~Zhuang, Y.~Zhuang, J.~E. Gonzalez, {\em et~al.}, ``Vicuna: An open-source chatbot impressing gpt-4 with 90\%* chatgpt quality.'' \url{https://vicuna.lmsys.org}, 2023.
\newblock Accessed: 2023-04-14.

\bibitem{DBLP:conf/eccv/LiuDZLZZYWHLCL24}
Y.~Liu, H.~Duan, Y.~Zhang, B.~Li, S.~Zhang, W.~Zhao, Y.~Yuan, J.~Wang, C.~He, Z.~Liu, K.~Chen, and D.~Lin, ``Mmbench: Is your multi-modal model an all-around player?,'' in {\em Computer Vision - {ECCV} 2024 - 18th European Conference, Milan, Italy, September 29-October 4, 2024, Proceedings, Part {VI}}, vol.~15064 of {\em Lecture Notes in Computer Science}, pp.~216--233, 2024.

\bibitem{DBLP:conf/acl/DaiDHSCW22}
D.~Dai, L.~Dong, Y.~Hao, Z.~Sui, B.~Chang, and F.~Wei, ``Knowledge neurons in pretrained transformers,'' in {\em Proceedings of the 60th Annual Meeting of the Association for Computational Linguistics (Volume 1: Long Papers), {ACL} 2022, Dublin, Ireland, May 22-27, 2022}, pp.~8493--8502, 2022.

\bibitem{KnowledgeEditor}
N.~D. Cao, W.~Aziz, and I.~Titov, ``Editing factual knowledge in language models,'' in {\em Proceedings of the 2021 Conference on Empirical Methods in Natural Language Processing, {EMNLP} 2021, Virtual Event / Punta Cana, Dominican Republic, 7-11 November, 2021}, pp.~6491--6506, 2021.

\bibitem{MEND}
E.~Mitchell, C.~Lin, A.~Bosselut, C.~Finn, and C.~D. Manning, ``Fast model editing at scale,'' in {\em The Tenth International Conference on Learning Representations, {ICLR} 2022, Virtual Event, April 25-29, 2022}, 2022.

\bibitem{DBLP:conf/iclr/TanZF24}
C.~Tan, G.~Zhang, and J.~Fu, ``Massive editing for large language models via meta learning,'' in {\em The Twelfth International Conference on Learning Representations, {ICLR} 2024, Vienna, Austria, May 7-11, 2024}, 2024.

\bibitem{DBLP:conf/acl/ZhangCL0HHXH24}
T.~Zhang, Q.~Chen, D.~Li, C.~Wang, X.~He, L.~Huang, H.~Xue, and J.~Huang, ``Dafnet: Dynamic auxiliary fusion for sequential model editing in large language models,'' in {\em Findings of the Association for Computational Linguistics, {ACL} 2024, Bangkok, Thailand and virtual meeting, August 11-16, 2024}, pp.~1588--1602, 2024.

\bibitem{T-Patcher}
Z.~Huang, Y.~Shen, X.~Zhang, J.~Zhou, W.~Rong, and Z.~Xiong, ``Transformer-patcher: One mistake worth one neuron,'' in {\em The Eleventh International Conference on Learning Representations, {ICLR} 2023, Kigali, Rwanda, May 1-5, 2023}, 2023.

\bibitem{LEMOE}
R.~Wang and P.~Li, ``Lemoe: Advanced mixture of experts adaptor for lifelong model editing of large language models,'' in {\em Proceedings of the 2024 Conference on Empirical Methods in Natural Language Processing, {EMNLP} 2024, Miami, FL, USA, November 12-16, 2024}, pp.~2551--2575, 2024.

\bibitem{SERAC}
E.~Mitchell, C.~Lin, A.~Bosselut, C.~D. Manning, and C.~Finn, ``Memory-based model editing at scale,'' in {\em International Conference on Machine Learning, {ICML} 2022, 17-23 July 2022, Baltimore, Maryland, {USA}}, vol.~162 of {\em Proceedings of Machine Learning Research}, pp.~15817--15831, 2022.

\bibitem{DBLP:conf/emnlp/MadaanTCY22}
A.~Madaan, N.~Tandon, P.~Clark, and Y.~Yang, ``Memory-assisted prompt editing to improve {GPT-3} after deployment,'' in {\em Proceedings of the 2022 Conference on Empirical Methods in Natural Language Processing, {EMNLP} 2022, Abu Dhabi, United Arab Emirates, December 7-11, 2022}, pp.~2833--2861, 2022.

\bibitem{DBLP:conf/acl/WangHB24}
W.~Wang, B.~Haddow, and A.~Birch, ``Retrieval-augmented multilingual knowledge editing,'' in {\em Proceedings of the 62nd Annual Meeting of the Association for Computational Linguistics (Volume 1: Long Papers), {ACL} 2024, Bangkok, Thailand, August 11-16, 2024}, pp.~335--354, 2024.

\bibitem{DBLP:conf/iclr/LepikhinLXCFHKS21}
D.~Lepikhin, H.~Lee, Y.~Xu, D.~Chen, O.~Firat, Y.~Huang, M.~Krikun, N.~Shazeer, and Z.~Chen, ``Gshard: Scaling giant models with conditional computation and automatic sharding,'' in {\em 9th International Conference on Learning Representations, {ICLR} 2021, Virtual Event, Austria, May 3-7, 2021}, 2021.

\bibitem{DBLP:journals/jmlr/FedusZS22}
W.~Fedus, B.~Zoph, and N.~Shazeer, ``Switch transformers: Scaling to trillion parameter models with simple and efficient sparsity,'' {\em J. Mach. Learn. Res.}, vol.~23, pp.~120:1--120:39, 2022.

\bibitem{DBLP:conf/icml/LewisBDGZ21}
M.~Lewis, S.~Bhosale, T.~Dettmers, N.~Goyal, and L.~Zettlemoyer, ``{BASE} layers: Simplifying training of large, sparse models,'' in {\em Proceedings of the 38th International Conference on Machine Learning, {ICML} 2021, 18-24 July 2021, Virtual Event}, vol.~139 of {\em Proceedings of Machine Learning Research}, pp.~6265--6274, 2021.

\bibitem{DBLP:conf/nips/RollerSSW21}
S.~Roller, S.~Sukhbaatar, A.~Szlam, and J.~Weston, ``Hash layers for large sparse models,'' in {\em Advances in Neural Information Processing Systems 34: Annual Conference on Neural Information Processing Systems 2021, NeurIPS 2021, December 6-14, 2021, virtual}, pp.~17555--17566, 2021.

\bibitem{DBLP:conf/nips/ZhouLLDHZDCLL22}
Y.~Zhou, T.~Lei, H.~Liu, N.~Du, Y.~Huang, V.~Y. Zhao, A.~M. Dai, Z.~Chen, Q.~V. Le, and J.~Laudon, ``Mixture-of-experts with expert choice routing,'' in {\em Advances in Neural Information Processing Systems 35: Annual Conference on Neural Information Processing Systems 2022, NeurIPS 2022, New Orleans, LA, USA, November 28 - December 9, 2022}, 2022.

\bibitem{DBLP:conf/icml/XueZFNZZ024}
F.~Xue, Z.~Zheng, Y.~Fu, J.~Ni, Z.~Zheng, W.~Zhou, and Y.~You, ``Openmoe: An early effort on open mixture-of-experts language models,'' in {\em Forty-first International Conference on Machine Learning, {ICML} 2024, Vienna, Austria, July 21-27, 2024}, 2024.

\bibitem{mistral2023mixtral}
M.~A. Team {\em et~al.}, ``Mixtral of experts: A high quality sparse mixture-of-experts,'' {\em Mistral AI Blog. Accessed: December}, vol.~18, p.~2023, 2023.

\bibitem{DBLP:journals/corr/abs-1807-03748}
A.~van~den Oord, Y.~Li, and O.~Vinyals, ``Representation learning with contrastive predictive coding,'' {\em CoRR}, vol.~abs/1807.03748, 2018.

\bibitem{DBLP:journals/corr/abs-2408-07413}
C.~Hu, P.~Cao, Y.~Chen, K.~Liu, and J.~Zhao, ``Knowledge in superposition: Unveiling the failures of lifelong knowledge editing for large language models,'' {\em CoRR}, vol.~abs/2408.07413, 2024.

\bibitem{DBLP:conf/acl/JawaharSS19}
G.~Jawahar, B.~Sagot, and D.~Seddah, ``What does {BERT} learn about the structure of language?,'' in {\em Proceedings of the 57th Conference of the Association for Computational Linguistics, {ACL} 2019, Florence, Italy, July 28- August 2, 2019, Volume 1: Long Papers}, pp.~3651--3657, 2019.

\bibitem{DBLP:conf/iccv/FongV17}
R.~C. Fong and A.~Vedaldi, ``Interpretable explanations of black boxes by meaningful perturbation,'' in {\em {IEEE} International Conference on Computer Vision, {ICCV} 2017, Venice, Italy, October 22-29, 2017}, pp.~3449--3457, 2017.

\bibitem{DBLP:journals/corr/ChenFLVGDZ15}
X.~Chen, H.~Fang, T.~Lin, R.~Vedantam, S.~Gupta, P.~Doll{\'{a}}r, and C.~L. Zitnick, ``Microsoft {COCO} captions: Data collection and evaluation server,'' {\em CoRR}, vol.~abs/1504.00325, 2015.

\bibitem{DBLP:conf/cvpr/RombachBLEO22}
R.~Rombach, A.~Blattmann, D.~Lorenz, P.~Esser, and B.~Ommer, ``High-resolution image synthesis with latent diffusion models,'' in {\em {IEEE/CVF} Conference on Computer Vision and Pattern Recognition, {CVPR} 2022, New Orleans, LA, USA, June 18-24, 2022}, pp.~10674--10685, 2022.

\bibitem{DBLP:conf/acl/DuQLDQY022}
Z.~Du, Y.~Qian, X.~Liu, M.~Ding, J.~Qiu, Z.~Yang, and J.~Tang, ``{GLM:} general language model pretraining with autoregressive blank infilling,'' in {\em Proceedings of the 60th Annual Meeting of the Association for Computational Linguistics (Volume 1: Long Papers), {ACL} 2022, Dublin, Ireland, May 22-27, 2022}, pp.~320--335, 2022.

\bibitem{DBLP:conf/cvpr/MarinoRFM19}
K.~Marino, M.~Rastegari, A.~Farhadi, and R.~Mottaghi, ``{OK-VQA:} {A} visual question answering benchmark requiring external knowledge,'' in {\em {IEEE} Conference on Computer Vision and Pattern Recognition, {CVPR} 2019, Long Beach, CA, USA, June 16-20, 2019}, pp.~3195--3204, 2019.

\bibitem{DBLP:conf/esws/LiuLGNOR19}
Y.~Liu, H.~Li, A.~Garc{\'{\i}}a{-}Dur{\'{a}}n, M.~Niepert, D.~O{\~{n}}oro{-}Rubio, and D.~S. Rosenblum, ``{MMKG:} multi-modal knowledge graphs,'' in {\em The Semantic Web - 16th International Conference, {ESWC} 2019, Portoro{\v{z}}, Slovenia, June 2-6, 2019, Proceedings}, vol.~11503 of {\em Lecture Notes in Computer Science}, pp.~459--474, 2019.

\bibitem{DBLP:conf/icml/RadfordKHRGASAM21}
A.~Radford, J.~W. Kim, C.~Hallacy, A.~Ramesh, G.~Goh, S.~Agarwal, G.~Sastry, A.~Askell, P.~Mishkin, J.~Clark, G.~Krueger, and I.~Sutskever, ``Learning transferable visual models from natural language supervision,'' in {\em Proceedings of the 38th International Conference on Machine Learning, {ICML} 2021, 18-24 July 2021, Virtual Event}, vol.~139 of {\em Proceedings of Machine Learning Research}, pp.~8748--8763, 2021.

\bibitem{DBLP:journals/corr/abs-2303-08774}
OpenAI, ``{GPT-4} technical report,'' {\em CoRR}, vol.~abs/2303.08774, 2023.

\bibitem{DBLP:conf/naacl/DevlinCLT19}
J.~Devlin, M.~Chang, K.~Lee, and K.~Toutanova, ``{BERT:} pre-training of deep bidirectional transformers for language understanding,'' in {\em Proceedings of the 2019 Conference of the North American Chapter of the Association for Computational Linguistics: Human Language Technologies, {NAACL-HLT} 2019, Minneapolis, MN, USA, June 2-7, 2019, Volume 1 (Long and Short Papers)} (J.~Burstein, C.~Doran, and T.~Solorio, eds.), pp.~4171--4186, Association for Computational Linguistics, 2019.

\bibitem{DBLP:journals/corr/abs-2205-01068}
S.~Zhang, S.~Roller, N.~Goyal, M.~Artetxe, M.~Chen, S.~Chen, C.~Dewan, M.~T. Diab, X.~Li, X.~V. Lin, T.~Mihaylov, M.~Ott, S.~Shleifer, K.~Shuster, D.~Simig, P.~S. Koura, A.~Sridhar, T.~Wang, and L.~Zettlemoyer, ``{OPT:} open pre-trained transformer language models,'' {\em CoRR}, vol.~abs/2205.01068, 2022.

\bibitem{DBLP:conf/emnlp/ReimersG19}
N.~Reimers and I.~Gurevych, ``Sentence-bert: Sentence embeddings using siamese bert-networks,'' in {\em Proceedings of the 2019 Conference on Empirical Methods in Natural Language Processing and the 9th International Joint Conference on Natural Language Processing, {EMNLP-IJCNLP} 2019, Hong Kong, China, November 3-7, 2019} (K.~Inui, J.~Jiang, V.~Ng, and X.~Wan, eds.), pp.~3980--3990, Association for Computational Linguistics, 2019.

\bibitem{DBLP:conf/nips/ZongMSSSJL024}
Z.~Zong, B.~Ma, D.~Shen, G.~Song, H.~Shao, D.~Jiang, H.~Li, and Y.~Liu, ``Mova: Adapting mixture of vision experts to multimodal context,'' in {\em Advances in Neural Information Processing Systems 38: Annual Conference on Neural Information Processing Systems 2024, NeurIPS 2024, Vancouver, BC, Canada, December 10 - 15, 2024} (A.~Globersons, L.~Mackey, D.~Belgrave, A.~Fan, U.~Paquet, J.~M. Tomczak, and C.~Zhang, eds.), 2024.

\end{thebibliography}
}


\clearpage
\setcounter{page}{1}
\maketitlesupplementary

\section{Details of Experimental Settings}
\label{Appendix_detail_exp_settings}

\subsection{Datasets:} 
\noindent\textbf{E-VQA} \cite{MMEdit} is a dataset designed to facilitate the correction of error-prone samples in VQA-v2 \cite{DBLP:conf/cvpr/GoyalKSBP17} for VLLMs, containing 6,345 training samples and 2,093 testing samples. The VQA task involves providing a VLLM with an image and a related question, requiring it to analyze both the visual content and the question to produce an accurate textual response.

\noindent\textbf{E-IC} \cite{MMEdit} is a dataset created for editing VLLMs to rectify errors in COCO Caption \cite{DBLP:journals/corr/ChenFLVGDZ15}, consisting of 2,849 training samples and 1,000 testing samples. The IC task requires the model to generate an accurate textual description for a given image, demanding a comprehensive understanding and articulation of the visual content.

Each sample in these two datasets includes an edit sample, along with two additional samples each for modal and text generality, and two samples each for modal and text locality. Generality samples are created by rephrasing images and queries using Stable Diffusion \cite{DBLP:conf/cvpr/RombachBLEO22} and ChatGLM \cite{DBLP:conf/acl/DuQLDQY022}, respectively. Meanwhile, locality samples are generated using unrelated images and queries from the OK-VQA \cite{DBLP:conf/cvpr/MarinoRFM19} and NQ \cite{MEND} datasets.

\noindent\textbf{VLKEB} \cite{VLKEB} is an editor evaluation dataset constructed based on MMKG (Multi-Modal Knowledge Graph) \cite{DBLP:conf/esws/LiuLGNOR19}. 
For each sample, it also includes an editing sample, along with two samples for modal and text generality, and two samples for modal and text locality.
It contains 5,000 training samples and 3,174 testing samples. 

Unlike the aforementioned datasets, VLKEB uses CLIP \cite{DBLP:conf/icml/RadfordKHRGASAM21} to retrieve semantically similar real-world images, which are subsequently verified manually. This process ensures higher image quality compared to datasets relying on image generation models, making VLKEB more representative of real-world scenarios.

\subsection{VLLM Backbones:} 
Below, we introduce the VLLM backbones used in our experiments. For the specific versions utilized, please refer to the corresponding footnotes.

\noindent\textbf{LLaVA}\footnote{\url{https://huggingface.co/liuhaotian/llava-v1.5-7b}} \cite{llava} is a Vision-Language Large Model (VLLM) that bridges the gap between visual and language understanding by leveraging GPT-4 \cite{DBLP:journals/corr/abs-2303-08774} to create an instruction-tuning dataset for vision-language pre-training. It achieves the transformation of visual features into the textual representation space through a simple yet effective two-layer MLP placed between the visual encoder and the language model LLaMA \cite{DBLP:journals/corr/abs-2302-13971}. This design allows LLaVA to excel at tasks such as visual question answering and instruction following with high efficiency and precision.

\noindent\textbf{BLIP2} \cite{BLIP2} introduces a visual query transformer, called Q-Former, which is trained through a two-stage pre-training process to compress visual features into fixed-length representations within the language space. This design bridges the representational gap between the frozen visual encoder \cite{DBLP:conf/icml/RadfordKHRGASAM21} and the frozen language model. BLIP2 comes in several variants, and in this paper, we follow \cite{MMEdit} to experiment with BLIP2-OPT\footnote{\url{https://huggingface.co/Salesforce/blip2-opt-2.7b}}. 
While LLaVa utilizes the full spectrum of visual features and preserves more fine-grained details, BLIP2 significantly reduces the length of visual representations, thereby improving the inference efficiency.

\noindent\textbf{MiniGPT-4}\footnote{\url{https://huggingface.co/Vision-CAIR/MiniGPT-4}} \cite{MiniGPT-4}, developed based on BLIP2, keeps the Q-Former and the language model Vicuna \cite{chiang2023vicuna} frozen, training only a linear layer positioned after the Q-Former to efficiently align visual features with Vicuna.

\begin{table*}[!tb]
    \footnotesize
    \centering
    \setlength{\tabcolsep}{4.5pt}
    \renewcommand{\arraystretch}{0.85}

\begin{tabular}{cccccc}
\toprule
\textbf{Editors}       & \textbf{Backbones} & \textbf{Edit Iterations} & \textbf{Optimizer} & \textbf{Learning Rate} & \textbf{Edit Layers}                        \\
\midrule
\multirow{3}{*}{FT-L}  & LLaVA-V1.5 (7B)    & 25                       & AdamW              & 1e-3                   & The last layer of the language transformer  \\
                       & BLIP2-OPT   (2.7B) & 25                       & AdamW              & 1e-3                   & The last layer of the language transformer  \\
                       & MiniGPT-4   (7B)   & 25                       & AdamW              & 1e-3                   & The last layer of the language transformer  \\
\midrule
\multirow{3}{*}{FT-M}  & LLaVA-V1.5 (7B)    & 25                       & AdamW              & 1e-3                   & The multi modal projector                   \\
                       & BLIP2-OPT   (2.7B) & 25                       & AdamW              & 1e-3                   & Qformer                                     \\
                       & MiniGPT-4   (7B)   & 25                       & AdamW              & 1e-3                   & Qformer                                     \\
\midrule
\multirow{3}{*}{TP}    & LLaVA-V1.5 (7B)    & 25                       & Adam               & 1e-2                   & The last layer of the language transformer  \\
                       & BLIP2-OPT   (2.7B) & 25                       & Adam               & 1e-2                   & The last layer of the language transformer  \\
                       & MiniGPT-4   (7B)   & 25                       & Adam               & 1e-2                   & The last layer of the language transformer  \\
\midrule
\multirow{3}{*}{LEMoE} & LLaVA-V1.5 (7B)    & 100                      & AdamW              & 2e-4                   & The 18-th layer of the language transformer \\
                       & BLIP2-OPT   (2.7B) & 100                      & AdamW              & 2e-4                   & The 18-th layer of the language transformer \\
                       & MiniGPT-4   (7B)   & 100                      & AdamW              & 2e-4                   & The 18-th layer of the language transformer \\
\bottomrule
\end{tabular}

    \caption{Editing details of direct editing editors.}
    \label{tab_exp_details_1}
\end{table*}

\begin{table*}[!tb]
    \footnotesize
    \centering
    \setlength{\tabcolsep}{4.5pt}
    \renewcommand{\arraystretch}{0.85}

\begin{tabular}{ccccc}
\toprule
\textbf{Editors}        & \textbf{Backbones} & \textbf{Optimizer} & \textbf{Learning Rate} & \textbf{Edit/Modified Layers}                       \\
\midrule
\multirow{3}{*}{MEND}   & LLaVA-V1.5 (7B)    & Adam               & 1e-4                   & layer 29,30,31 of the language transformer \\
                        & BLIP2-OPT   (2.7B) & Adam               & 1e-4                   & layer 29,30,31 of the language transformer \\
                        & MiniGPT-4   (7B)   & Adam               & 1e-4                   & layer 29,30,31 of the language transformer \\
\midrule
\multirow{3}{*}{SERAC}  & LLaVA-V1.5 (7B)    & Adam               & 1e-4                   & -                                          \\
                        & BLIP2-OPT   (2.7B) & Adam               & 1e-4                   & -                                          \\
                        & MiniGPT-4   (7B)   & Adam               & 1e-4                   & -                                          \\
\midrule
\multirow{3}{*}{LTE}    & LLaVA-V1.5 (7B)    & Adam               & 5e-6                   & The whole language transformer                                          \\
                        & BLIP2-OPT   (2.7B) & Adam               & 5e-6                   & The whole language transformer                                          \\
                        & MiniGPT-4   (7B)   & Adam               & 5e-6                   & The whole language transformer                                          \\
\midrule
\multirow{3}{*}{RECIPE} & LLaVA-V1.5 (7B)    & Adam               & 1e-5                   & -                                          \\
                        & BLIP2-OPT   (2.7B) & Adam               & 1e-5                   & -                                          \\
                        & MiniGPT-4   (7B)   & Adam               & 1e-5                   & -                                         \\
\bottomrule
\end{tabular}

    \caption{Editing details of training-based editing methods.}
    \label{tab_exp_details_2}
\end{table*}

\subsection{Baseline Editors:} 
\noindent\textbf{FT} (\textbf{F}ine-\textbf{T}uning) comprises two variants: FT-L and FT-M. According to \cite{MMEdit}, FT-L involves fine-tuning the final layer of the language transformer within the VLLM, whereas FT-M focuses on fine-tuning the visual encoder module of the VLLM for a given edit sample.

\noindent\textbf{MEND} (\textbf{M}odel \textbf{E}ditor \textbf{N}etworks with \textbf{D}ecomposition) \cite{MEND} trains a set of small MLP hyper-networks to achieve efficient model editing. These hyper-networks generate the FFN matrix parameter offsets by inputting the decomposed back-propagation gradients of these matrices on edit samples. Through editing-specific training, these hyper-networks can produce matrix offsets that enable LLMs to satisfy editing requirements.

\noindent\textbf{SERAC} (\textbf{S}emi-parametric \textbf{E}diting with a \textbf{R}etrieval-\textbf{A}ugmented \textbf{C}ounterfactual) \cite{SERAC} is a memory-based method. It trains a scope classifier and a small counterfactual model while storing edit samples in memory. The classifier determines whether subsequent inputs are related to the edit samples. If they are related, the inputs are sent to the counterfactual model for response modification; otherwise, they are passed to the original model to generate responses. In our experiments, we follow the setup of MMEdit \cite{MMEdit}, where BERT\footnote{\url{https://huggingface.co/google-bert/bert-base-cased}} \cite{DBLP:conf/naacl/DevlinCLT19} and OPT-125M\footnote{\url{https://huggingface.co/facebook/opt-125m}} \cite{DBLP:journals/corr/abs-2205-01068} are separately trained as the scope classifier and the counterfactual model.

\noindent\textbf{TP} (\textbf{T}ransformer \textbf{P}atcher) \cite{T-Patcher} assumes that each neuron in the LLM can carry a piece of knowledge. Therefore, for every new piece of edit sample, TP inserts and trains a new neuron in the final layer of the FFN within the LLM to accommodate the edit sample.

\noindent\textbf{LTE} (\textbf{L}earning \textbf{T}o \textbf{E}dit) \cite{LTE} performs supervised fine-tuning of LLMs on edit samples, enabling the model to adapt its response when an edit sample is provided as a contextual prefix. To address sequential editing scenarios, it stores edit samples in a retrieval database and uses a pre-existing text retriever \cite{DBLP:conf/emnlp/ReimersG19} to retrieve edit samples relevant to the input.

\noindent\textbf{RECIPE} (\textbf{RE}trieval-augmented \textbf{C}ont\textbf{I}nue \textbf{P}rompt l\textbf{E}arning) \cite{RECIPE} aims to explore short editing prefixes to achieve lifelong editing for LLMs. By training an editing prompt generator, edit samples are converted into short continuous prompts, which serve as prefix contexts for LLMs to adapt their responses. Additionally, RECIPE trains an editing retriever to retrieve relevant edited samples based on the input.

\noindent\textbf{LEMoE} (\textbf{L}ifelong \textbf{E}diting with \textbf{M}ixture \textbf{o}f \textbf{E}xperts) \cite{LEMOE} achieves lifelong editing for LLMs by inserting and maintaining a MoE in the final layer of the LLM. In lifelong editing, whenever a batch of editing requirements arises, an editing expert is initialized and trained to adapt the model's responses for that specific batch of edit samples. During the training of a new editing expert, previously trained experts also participate in the back-propagation process, but their parameters remain frozen. Additionally, a key vector is trained alongside the expert to route the input to the relevant experts.

\subsection{Model Settings and Training Details}
\noindent\textbf{LiveEdit:} LiveEdit adopts the same hyperparameters for all backbones. After hyperparameter tuning and balancing computational resources with editing performance, we set the module dimension $d_m = 1024$, the rank of the editing expert $r = 4$, and the feature extraction control parameter $k = 4$. 
For the index of editing layers $l_e$, based on the attribution analysis from \cite{DBLP:journals/corr/abs-2408-09916} and our experiments in Figure \ref{fig_module_put_layer}, we set $l_e = 21$. The learning rate $\eta$ is set to $1\text{e-4}$, the training batch size $B = 8$, and the maximum number of training iterations is 200K. 
If the loss stops decreasing, we terminate the training process early.
A checkpoint is saved every 500 iterations, and the one with the lowest loss is selected for lifelong editing evaluation. The training process takes approximately 2 days on a single NVIDIA A800 GPU.

\noindent\textbf{Other Baselines:} The hyperparameters of TP \cite{T-Patcher}, LTE \cite{LTE}, and LEMoE \cite{LEMOE} are adopted from their respective papers. They are slightly adjusted to accommodate VLLM editing, such as mapping visual representations to the language representation space via a trainable MLP module. For the other baselines, we follow \cite{MMEdit}. 
For the experimental details of direct editing methods, please refer to Table \ref{tab_exp_details_1}.
For training-based editing methods, we also set the maximum number of iterations to 200K. Training details are provided in Table \ref{tab_exp_details_2}.

\begin{table*}[!tb]
    \scriptsize
    \centering
    \setlength{\tabcolsep}{0.7pt}
    \renewcommand{\arraystretch}{0.85}

\begin{tabular}{ccccccccccccccccccccc}
\toprule
\multirow{2}{*}{\textbf{Baseline}}  & \multirow{2}{*}{\textbf{\# Edit}} & \multirow{2}{*}{\textbf{Editors}} & \multicolumn{6}{c}{\textbf{E-VQA}}  & \multicolumn{6}{c}{\textbf{VLKEB}}  & \multicolumn{6}{c}{\textbf{E-IC}}  \textbf{}  \\
  & & & \textbf{Rel.}  & \textbf{T-Gen.} & \textbf{M-Gen.} & \textbf{T-Loc.} & \textbf{M-Loc.} & \textbf{Average} & \textbf{Rel.}  & \textbf{T-Gen.} & \textbf{M-Gen.} & \textbf{T-Loc.} & \textbf{M-Loc.} & \textbf{Average} & \textbf{Rel.}  & \textbf{T-Gen.} & \textbf{M-Gen.} & \textbf{T-Loc.} & \textbf{M-Loc.} & \textbf{Average}  \textbf{}  \\
\midrule
\multirow{41}{*}{\begin{tabular}[c]{@{}c@{}}LLaVA\\  (7B)\end{tabular}}  & \multirow{9}{*}{1}  & FT-L  & 93.88  & 87.98 & 80.25 & 99.61 & 94.78 & 91.30 $_{(\pm0.42)}$ & 94.29  & 87.00 & 92.22 & 91.16 & 91.37 & 91.21 $_{(\pm1.09)}$ & 73.48  & 72.98 & 65.79 & 99.28 & 99.06 & 82.12 $_{(\pm0.82)}$ \\
  & & FT-M  & 87.29  & 76.11 & 53.23 & 100.00  & 96.95 & 82.72 $_{(\pm1.05)}$ & 76.31  & 65.57 & 59.43 & 100.00  & 92.35 & 78.73 $_{(\pm0.76)}$ & 56.19  & 56.55 & 49.94 & 100.00  & 100.00  & 72.54 $_{(\pm0.85)}$ \\
  & & MEND  & 91.23  & 90.05 & 91.29 & 91.02 & 90.22 & 90.76 $_{(\pm0.64)}$ & 92.13  & 91.28 & 90.22 & 89.19 & 90.13 & 90.59 $_{(\pm1.24)}$ & 92.82  & \textbf{91.81}  & 90.59 & 96.38 & 93.69 & \textbf{93.06}  $_{(\pm1.40)}$ \\
  & & SERAC & 89.33  & 83.72 & 84.97 & 82.05 & 23.78 & 72.77 $_{(\pm0.36)}$ & 89.77  & 89.11 & 87.92 & 66.68 & 14.20 & 69.54 $_{(\pm0.83)}$ & 88.18  & 81.03 & 85.61 & 84.01 & 28.58 & 73.48 $_{(\pm1.19)}$ \\
  & & TP  & 35.95  & 36.12 & 28.65 & 93.87 & 97.61 & 58.44 $_{(\pm0.33)}$ & 50.77  & 55.70 & 51.65 & 87.93 & 90.43 & 67.30 $_{(\pm0.29)}$ & 57.63  & 59.23 & 55.34 & 60.90 & 88.00 & 64.22 $_{(\pm0.49)}$ \\
  & & LTE & 94.16  & 93.57 & \textbf{93.59}  & 94.08 & 86.26 & 92.33 $_{(\pm1.56)}$ & 94.42  & 93.57 & 93.22 & 86.84 & 79.69 & 89.55 $_{(\pm1.41)}$ & 93.35  & 91.30 & \textbf{92.77}  & 95.77 & 91.98 & 93.03 $_{(\pm0.55)}$ \\
  & & RECIPE  & 91.37  & 86.51 & 87.73 & 94.27 & 88.88 & 89.75 $_{(\pm1.13)}$ & 92.67  & 92.35 & 91.01 & 89.67 & 82.85 & 89.71 $_{(\pm0.57)}$ & 84.45  & 76.97 & 81.57 & 96.53 & 96.37 & 87.18 $_{(\pm1.08)}$ \\
  & & LEMoE & 93.60  & 92.77 & 89.99 & 99.28 & 96.98 & 94.52 $_{(\pm1.09)}$ & 94.85  & 93.09 & 91.67 & 87.03 & 87.88 & 90.90 $_{(\pm0.29)}$ & \textbf{93.80} & 91.42 & 90.61 & 95.14 & 93.00 & 92.79 $_{(\pm0.41)}$ \\
  & & LiveEdit & \textbf{94.28} & \textbf{94.51}  & 88.01 & \textbf{100.00} & \textbf{100.00} & \textbf{95.36}  $_{(\pm0.57)}$ & \textbf{96.43} & \textbf{95.22}  & \textbf{93.72}  & \textbf{100.00} & \textbf{100.00} & \textbf{97.08}  $_{(\pm0.62)}$ & 82.16  & 81.01 & 78.27 & \textbf{100.00} & \textbf{100.00} & 88.29 $_{(\pm1.42)}$ \\
\cmidrule{2-21}  
  & \multirow{9}{*}{10} & FT-L  & 90.57  & 84.14 & 73.21 & 95.56 & 81.50 & 85.00 $_{(\pm1.07)}$ & 88.05  & 85.32 & 85.23 & 74.53 & 85.74 & 83.77 $_{(\pm1.22)}$ & 68.74  & 67.05 & 60.95 & 97.05 & 91.20 & 77.00 $_{(\pm0.80)}$ \\
  & & FT-M  & 84.90  & 73.53 & 49.99 & 100.00  & 55.98 & 72.88 $_{(\pm0.63)}$ & 68.63  & 57.57 & 56.56 & 100.00  & 82.99 & 73.15 $_{(\pm0.23)}$ & 57.23  & 55.92 & 50.09 & 100.00  & 90.31 & 70.71 $_{(\pm1.18)}$ \\
  & & MEND  & 3.58 & 3.55  & 3.53  & 2.10  & 1.26  & 2.80  $_{(\pm0.02)}$ & 0.18 & 0.24  & 0.05  & 0.03  & 0.19  & 0.14  $_{(\pm0.00)}$ & 66.49  & 64.77 & 58.42 & 87.25 & 86.15 & 72.62 $_{(\pm1.13)}$ \\
  & & SERAC & 88.09  & 83.40 & 83.57 & 64.91 & 15.50 & 67.10 $_{(\pm0.92)}$ & 81.55  & 74.49 & 80.24 & 54.71 & 13.15 & 60.83 $_{(\pm0.98)}$ & 56.57  & 56.88 & 52.62 & 59.96 & 14.66 & 48.14 $_{(\pm0.40)}$ \\
  & & TP  & 32.71  & 31.23 & 28.58 & 75.10 & 91.17 & 51.76 $_{(\pm0.60)}$ & 44.56  & 47.52 & 45.36 & 52.21 & 66.61 & 51.25 $_{(\pm0.69)}$ & 45.28  & 47.25 & 42.78 & 19.74 & 59.59 & 42.93 $_{(\pm0.38)}$ \\
  & & LTE & 92.83  & 91.41 & \textbf{90.82}  & 86.38 & 85.52 & 89.39 $_{(\pm0.34)}$ & 90.06  & 81.52 & 88.11 & 83.40 & 81.48 & 84.91 $_{(\pm0.78)}$ & 52.16  & 55.70 & 48.39 & 90.74 & 89.09 & 67.21 $_{(\pm0.89)}$ \\
  & & RECIPE  & 90.22  & 85.92 & 86.24 & 90.34 & 88.11 & 88.17 $_{(\pm1.48)}$ & 83.92  & 76.23 & 82.84 & 86.33 & 83.69 & 82.60 $_{(\pm0.72)}$ & 56.00  & 56.19 & 52.14 & 91.80 & 95.31 & 70.29 $_{(\pm0.98)}$ \\
  & & LEMoE & 91.95  & 86.54 & 79.82 & 85.19 & 49.81 & 78.66 $_{(\pm1.03)}$ & 91.55  & 84.58 & 81.03 & 67.19 & 72.81 & 79.43 $_{(\pm0.52)}$ & \textbf{89.00} & \textbf{85.23}  & \textbf{83.24}  & 86.39 & 82.86 & 85.34 $_{(\pm0.91)}$ \\
  & & LiveEdit & \textbf{93.79} & \textbf{93.21}  & 86.42 & \textbf{100.00} & \textbf{100.00} & \textbf{94.68}  $_{(\pm1.03)}$ & \textbf{95.54} & \textbf{94.52}  & \textbf{91.25}  & \textbf{100.00} & \textbf{100.00} & \textbf{96.26}  $_{(\pm0.33)}$ & 81.93  & 80.80 & 75.55 & \textbf{100.00} & \textbf{100.00} & \textbf{87.66}  $_{(\pm0.49)}$ \\
\cmidrule{2-21}  
  & \multirow{9}{*}{100}  & FT-L  & 79.67  & 70.05 & 64.07 & 83.47 & 54.44 & 70.34 $_{(\pm0.95)}$ & 75.41  & 73.67 & 74.16 & 70.01 & 82.05 & 75.06 $_{(\pm1.17)}$ & 65.08  & 60.90 & 58.46 & 86.82 & 89.19 & 72.09 $_{(\pm0.65)}$ \\
  & & FT-M  & 82.90  & 72.83 & 47.26 & 100.00  & 43.39 & 69.28 $_{(\pm0.43)}$ & 60.60  & 59.79 & 56.29 & 100.00  & 68.07 & 68.95 $_{(\pm0.78)}$ & 59.41  & 55.44 & 52.16 & 100.00  & 72.93 & 67.99 $_{(\pm0.43)}$ \\
  & & MEND  & 2.22 & 2.20  & 2.21  & 0.21  & 0.62  & 1.49  $_{(\pm0.02)}$ & 0.56 & 0.58  & 0.66  & 0.18  & 0.07  & 0.41  $_{(\pm0.00)}$ & 56.83  & 56.89 & 53.07 & 87.63 & 84.64 & 67.81 $_{(\pm0.93)}$ \\
  & & SERAC & 88.08  & 81.53 & 82.48 & 62.13 & 12.90 & 65.42 $_{(\pm0.32)}$ & 72.25  & 62.43 & 70.68 & 53.73 & 13.69 & 54.56 $_{(\pm0.49)}$ & 53.35  & 53.70 & 49.41 & 48.04 & 17.25 & 44.35 $_{(\pm0.50)}$ \\
  & & TP  & 29.37  & 28.72 & 24.66 & 14.64 & 45.01 & 28.48 $_{(\pm0.33)}$ & 19.71  & 20.07 & 19.36 & 11.40 & 24.05 & 18.92 $_{(\pm0.08)}$ & 22.88  & 25.81 & 20.90 & 3.59  & 14.87 & 17.61 $_{(\pm0.06)}$ \\
  & & LTE & 88.92  & 87.89 & 85.72 & 84.34 & 81.60 & 85.69 $_{(\pm0.37)}$ & 80.27  & 64.25 & 80.13 & 81.62 & 79.11 & 77.08 $_{(\pm0.85)}$ & 48.37  & 49.63 & 45.08 & 87.66 & 87.62 & 63.67 $_{(\pm0.60)}$ \\
  & & RECIPE  & 89.86  & 83.32 & 84.82 & 87.37 & 85.08 & 86.09 $_{(\pm0.55)}$ & 73.97  & 63.73 & 72.69 & 86.20 & 82.59 & 75.84 $_{(\pm0.58)}$ & 53.23  & 53.75 & 49.36 & 87.64 & 95.52 & 67.90 $_{(\pm0.36)}$ \\
  & & LEMoE & 42.41  & 36.60 & 34.33 & 78.57 & 53.28 & 49.04 $_{(\pm0.45)}$ & 83.07  & 75.36 & 71.72 & 54.09 & 49.68 & 66.78 $_{(\pm0.60)}$ & 55.36  & 53.14 & 51.12 & 87.77 & 81.35 & 65.75 $_{(\pm1.02)}$ \\
  & & LiveEdit & \textbf{93.54} & \textbf{92.34}  & \textbf{85.89}  & \textbf{100.00} & \textbf{99.31}  & \textbf{94.21}  $_{(\pm0.34)}$ & \textbf{94.56} & \textbf{90.65}  & \textbf{89.56}  & \textbf{100.00} & \textbf{100.00} & \textbf{94.95}  $_{(\pm1.34)}$ & \textbf{80.81} & \textbf{78.77}  & \textbf{63.52}  & \textbf{100.00} & \textbf{100.00} & \textbf{84.62}  $_{(\pm0.62)}$ \\
\cmidrule{2-21}  
  & \multirow{9}{*}{1000} & FT-L  & 71.39  & 59.83 & 57.41 & 55.55 & 48.99 & 58.63 $_{(\pm0.17)}$ & 68.14  & 66.38 & 66.98 & 65.61 & 75.35 & 68.49 $_{(\pm0.32)}$ & 59.78  & 54.99 & 54.17 & 65.37 & 78.96 & 62.65 $_{(\pm0.43)}$ \\
  & & FT-M  & 69.57  & 56.34 & 44.07 & 100.00  & 41.47 & 62.29 $_{(\pm0.40)}$ & 53.41  & 48.80 & 43.16 & 100.00  & 57.03 & 60.48 $_{(\pm0.50)}$ & 49.21  & 47.75 & 43.81 & 100.00  & 35.14 & 55.18 $_{(\pm0.88)}$ \\
  & & MEND  & 0.04 & 0.05  & 0.05  & 0.08  & 0.09  & 0.06  $_{(\pm0.00)}$ & 0.03 & 0.05  & 0.07  & 0.06  & 0.08  & 0.06  $_{(\pm0.00)}$ & 54.39  & 54.14 & 50.99 & 83.87 & 80.60 & 64.80 $_{(\pm0.35)}$ \\
  & & SERAC & 85.57  & 75.58 & 82.01 & 62.46 & 15.69 & 64.26 $_{(\pm0.37)}$ & 60.93  & 56.49 & 60.06 & 52.94 & 15.04 & 49.09 $_{(\pm0.36)}$ & 52.93  & 53.44 & 49.01 & 49.91 & 16.65 & 44.39 $_{(\pm0.73)}$ \\
  & & TP  & 16.56  & 16.80 & 15.65 & 7.28  & 15.60 & 14.38 $_{(\pm0.14)}$ & 5.46 & 4.81  & 5.51  & 2.77  & 7.19  & 5.15  $_{(\pm0.07)}$ & 10.28  & 13.14 & 9.75  & 1.71  & 4.45  & 7.87  $_{(\pm0.13)}$ \\
  & & LTE & 83.93  & 82.55 & 81.34 & 83.97 & 73.09 & 80.98 $_{(\pm1.36)}$ & 64.51  & 56.26 & 64.80 & 80.85 & 76.52 & 68.59 $_{(\pm0.60)}$ & 48.83  & 49.96 & 45.68 & 85.17 & 86.41 & 63.21 $_{(\pm0.67)}$ \\
  & & RECIPE  & 87.00  & 76.81 & 83.09 & 86.95 & 87.03 & 84.18 $_{(\pm0.80)}$ & 62.00  & 56.84 & 61.50 & 85.37 & 82.07 & 69.56 $_{(\pm0.31)}$ & 53.11  & 53.48 & 48.99 & 87.93 & 94.84 & 67.67 $_{(\pm1.11)}$ \\
  & & LEMoE & 30.80  & 25.75 & 24.32 & 71.45 & 46.23 & 39.71 $_{(\pm0.23)}$ & 67.97  & 61.07 & 58.16 & 48.48 & 44.06 & 55.95 $_{(\pm0.36)}$ & 34.50  & 31.38 & 28.14 & 82.09 & 75.88 & 50.40 $_{(\pm0.26)}$ \\
  & & LiveEdit & \textbf{92.93} & \textbf{90.16}  & \textbf{84.30}  & \textbf{100.00} & \textbf{96.43}  & \textbf{92.76}  $_{(\pm0.20)}$ & \textbf{92.22} & \textbf{83.97}  & \textbf{82.75}  & \textbf{100.00} & \textbf{100.00} & \textbf{91.79}  $_{(\pm0.55)}$ & \textbf{72.80} & \textbf{69.95}  & \textbf{57.05}  & \textbf{100.00} & \textbf{99.79}  & \textbf{79.92}  $_{(\pm0.72)}$ \\
\midrule
\multirow{41}{*}{\begin{tabular}[c]{@{}c@{}}BLIP2\\(2.7B)\end{tabular}} & \multirow{9}{*}{1}  & FT-L  & 52.86  & 48.80 & 32.94 & 98.24 & 94.27 & 65.42 $_{(\pm0.69)}$ & 54.31  & 54.27 & 54.08 & 98.40 & 94.37 & 71.09 $_{(\pm1.05)}$ & 45.02  & 44.47 & 40.72 & 99.02 & 98.27 & 65.50 $_{(\pm0.49)}$ \\
  & & FT-M  & 91.70  & 87.24 & 33.30 & 100.00  & 85.22 & 79.49 $_{(\pm0.72)}$ & 92.64  & 80.97 & 63.62 & 100.00  & 83.02 & 84.05 $_{(\pm0.70)}$ & 67.14  & 61.76 & 43.34 & 100.00  & 96.76 & 73.80 $_{(\pm0.16)}$ \\
  & & MEND  & 93.13  & 92.76 & 93.07 & 92.00 & 75.81 & 89.35 $_{(\pm0.93)}$ & 94.91  & 93.81 & 93.84 & 94.98 & 86.54 & 92.82 $_{(\pm0.82)}$ & \textbf{94.96} & \textbf{92.45}  & \textbf{92.33}  & 94.95 & 88.86 & 92.71 $_{(\pm1.45)}$ \\
  & & SERAC & 88.39  & 84.50 & 84.25 & 85.82 & 26.00 & 73.79 $_{(\pm1.01)}$ & 87.95  & 84.67 & 85.20 & 68.10 & 17.75 & 68.73 $_{(\pm0.97)}$ & 88.71  & 83.81 & 84.38 & 84.28 & 24.70 & 73.18 $_{(\pm0.25)}$ \\
  & & TP  & 70.14  & 65.80 & 53.05 & 98.11 & 85.33 & 74.49 $_{(\pm0.38)}$ & 50.98  & 49.47 & 50.88 & 94.76 & 78.57 & 64.93 $_{(\pm1.02)}$ & 49.65  & 48.58 & 46.02 & 93.69 & 78.95 & 63.38 $_{(\pm1.00)}$ \\
  & & LTE & 95.74  & 93.86 & 86.90 & 97.93 & 87.97 & 92.48 $_{(\pm0.70)}$ & 94.13  & 91.93 & 92.23 & 93.89 & 92.27 & 92.89 $_{(\pm1.01)}$ & 92.58  & 91.94 & 90.90 & 97.80 & 91.42 & \textbf{92.93}  $_{(\pm1.37)}$ \\
  & & RECIPE  & 89.42  & 86.24 & 87.53 & 99.87 & 89.16 & 90.45 $_{(\pm1.46)}$ & 92.38  & 89.74 & 89.17 & 97.13 & 94.46 & 92.58 $_{(\pm1.16)}$ & 85.20  & 81.44 & 82.71 & 100.00  & 94.59 & 88.79 $_{(\pm1.23)}$ \\
  & & LEMoE & 93.56  & 92.23 & 91.40 & 98.50 & 85.21 & 92.18 $_{(\pm0.73)}$ & 94.59  & 93.14 & 92.37 & 94.53 & 61.53 & 87.23 $_{(\pm0.34)}$ & 93.07  & 91.37 & 83.28 & 94.45 & 60.44 & 84.52 $_{(\pm0.38)}$ \\
  & & LiveEdit & \textbf{96.67} & \textbf{94.20}  & \textbf{93.82}  & \textbf{100.00} & \textbf{100.00} & \textbf{96.94}  $_{(\pm1.32)}$ & \textbf{98.77} & \textbf{98.08}  & \textbf{94.89}  & \textbf{100.00} & \textbf{100.00} & \textbf{98.35}  $_{(\pm1.58)}$ & 80.60  & 80.12 & 76.88 & \textbf{100.00} & \textbf{100.00} & 87.52 $_{(\pm0.31)}$ \\
\cmidrule{2-21}  
  & \multirow{9}{*}{10} & FT-L  & 53.77  & 48.91 & 37.66 & 97.68 & 80.40 & 63.68 $_{(\pm0.88)}$ & 55.81  & 55.55 & 55.29 & 93.37 & 75.45 & 67.09 $_{(\pm0.71)}$ & 46.20  & 44.07 & 40.54 & 98.37 & 92.59 & 64.35 $_{(\pm0.75)}$ \\
  & & FT-M  & 85.27  & 81.09 & 34.60 & 100.00  & 49.00 & 69.99 $_{(\pm0.15)}$ & 92.48  & 86.02 & 71.48 & 100.00  & 60.34 & 82.06 $_{(\pm0.47)}$ & 63.79  & 56.88 & 47.75 & 100.00  & 52.41 & 64.17 $_{(\pm0.42)}$ \\
  & & MEND  & 40.40  & 32.21 & 27.73 & 87.92 & 67.74 & 51.20 $_{(\pm0.39)}$ & 54.47  & 52.59 & 53.61 & 91.96 & 84.58 & 67.44 $_{(\pm0.16)}$ & 7.15 & 7.48  & 7.01  & 19.74 & 21.70 & 12.62 $_{(\pm0.05)}$ \\
  & & SERAC & 86.12  & 82.24 & 82.52 & 70.24 & 16.06 & 67.43 $_{(\pm1.10)}$ & 78.43  & 67.72 & 75.12 & 56.21 & 14.54 & 58.40 $_{(\pm0.73)}$ & 47.69  & 46.94 & 43.46 & 65.29 & 16.65 & 44.01 $_{(\pm0.15)}$ \\
  & & TP  & 55.86  & 52.39 & 43.00 & 91.26 & 63.62 & 61.22 $_{(\pm0.54)}$ & 50.19  & 51.61 & 50.13 & 89.26 & 68.48 & 61.93 $_{(\pm0.16)}$ & 44.57  & 44.88 & 39.82 & 65.84 & 51.55 & 49.33 $_{(\pm0.37)}$ \\
  & & LTE & 94.96  & 92.43 & 85.23 & 94.19 & 87.01 & 90.76 $_{(\pm1.19)}$ & 92.02  & 80.25 & 90.30 & 94.66 & 90.39 & 89.52 $_{(\pm1.47)}$ & 47.56  & 50.73 & 44.64 & 97.05 & 92.90 & 66.57 $_{(\pm0.78)}$ \\
  & & RECIPE  & 88.27  & 85.37 & 84.80 & 97.78 & 88.49 & 88.94 $_{(\pm0.76)}$ & 81.39  & 70.31 & 79.19 & 94.40 & 95.59 & 84.18 $_{(\pm0.74)}$ & 47.26  & 46.30 & 42.83 & 100.00  & 95.38 & 66.35 $_{(\pm0.76)}$ \\
  & & LEMoE & 91.59  & 85.92 & 86.82 & 83.64 & 29.71 & 75.54 $_{(\pm0.86)}$ & 91.40  & 90.71 & 89.12 & 58.20 & 49.49 & 75.79 $_{(\pm0.48)}$ & \textbf{87.45} & \textbf{84.54}  & 71.74 & 85.98 & 71.23 & 80.19 $_{(\pm0.60)}$ \\
  & & LiveEdit & \textbf{95.65} & \textbf{93.46}  & \textbf{90.98}  & \textbf{100.00} & \textbf{100.00} & \textbf{96.02}  $_{(\pm0.93)}$ & \textbf{98.57} & \textbf{97.92}  & \textbf{94.63}  & \textbf{100.00} & \textbf{100.00} & \textbf{98.22}  $_{(\pm0.48)}$ & 80.65  & 79.38 & \textbf{75.66}  & \textbf{100.00} & \textbf{100.00} & \textbf{87.14}  $_{(\pm0.96)}$ \\
\cmidrule{2-21}  
  & \multirow{9}{*}{100}  & FT-L  & 52.47  & 43.68 & 39.43 & 91.13 & 47.72 & 54.88 $_{(\pm0.83)}$ & 57.26  & 56.23 & 56.73 & 86.19 & 68.87 & 65.06 $_{(\pm0.15)}$ & 50.81  & 48.70 & 42.79 & 90.12 & 52.09 & 56.90 $_{(\pm0.89)}$ \\
  & & FT-M  & 48.04  & 40.71 & 29.86 & 100.00  & 33.25 & 50.37 $_{(\pm0.43)}$ & 56.49  & 58.16 & 53.13 & 100.00  & 43.52 & 62.26 $_{(\pm0.91)}$ & 56.23  & 53.18 & 43.37 & 100.00  & 31.59 & 56.87 $_{(\pm0.44)}$ \\
  & & MEND  & 17.69  & 16.39 & 18.31 & 91.52 & 67.94 & 42.37 $_{(\pm0.45)}$ & 39.91  & 40.96 & 40.37 & 92.32 & 84.32 & 59.58 $_{(\pm0.19)}$ & 7.97 & 8.19  & 8.04  & 20.15 & 23.23 & 13.52 $_{(\pm0.13)}$ \\
  & & SERAC & 86.98  & 81.27 & 80.97 & 71.04 & 15.55 & 67.17 $_{(\pm0.60)}$ & 66.73  & 53.69 & 64.37 & 59.18 & 17.86 & 52.37 $_{(\pm0.70)}$ & 43.55  & 42.16 & 39.09 & 57.77 & 15.84 & 39.68 $_{(\pm0.50)}$ \\
  & & TP  & 44.26  & 38.24 & 33.28 & 43.81 & 38.54 & 39.63 $_{(\pm0.10)}$ & 46.61  & 48.12 & 47.22 & 64.25 & 43.21 & 49.88 $_{(\pm0.49)}$ & 37.51  & 37.62 & 33.80 & 10.26 & 20.84 & 28.00 $_{(\pm0.47)}$ \\
  & & LTE & 92.78  & 90.36 & 83.88 & 94.33 & 81.79 & 88.63 $_{(\pm0.48)}$ & 78.41  & 60.70 & 78.55 & 94.34 & 90.13 & 80.42 $_{(\pm0.72)}$ & 43.03  & 42.52 & 39.63 & 96.20 & 91.95 & 62.67 $_{(\pm0.81)}$ \\
  & & RECIPE  & 88.12  & 82.49 & 83.10 & 98.97 & 86.53 & 87.84 $_{(\pm1.06)}$ & 68.79  & 54.96 & 67.20 & 94.61 & 97.12 & 76.54 $_{(\pm0.77)}$ & 43.40  & 42.19 & 39.02 & 98.40 & 95.55 & 63.71 $_{(\pm0.86)}$ \\
  & & LEMoE & 29.61  & 21.70 & 27.05 & 79.48 & 32.61 & 38.09 $_{(\pm0.11)}$ & 44.13  & 44.78 & 42.55 & 58.13 & 51.76 & 48.27 $_{(\pm0.71)}$ & 57.13  & 51.73 & 46.73 & 91.72 & 58.06 & 61.08 $_{(\pm0.34)}$ \\
  & & LiveEdit & \textbf{95.25} & \textbf{92.94}  & \textbf{85.48}  & \textbf{100.00} & \textbf{99.76}  & \textbf{94.69}  $_{(\pm0.50)}$ & \textbf{98.20} & \textbf{97.67}  & \textbf{93.96}  & \textbf{100.00} & \textbf{100.00} & \textbf{97.97}  $_{(\pm0.34)}$ & \textbf{79.18} & \textbf{77.44}  & \textbf{74.12}  & \textbf{100.00} & \textbf{100.00} & \textbf{86.15}  $_{(\pm0.33)}$ \\
\cmidrule{2-21}  
  & \multirow{9}{*}{1000} & FT-L  & 45.10  & 34.62 & 35.42 & 48.42 & 41.24 & 40.96 $_{(\pm0.29)}$ & 55.39  & 54.34 & 53.87 & 50.80 & 54.00 & 53.68 $_{(\pm0.80)}$ & 53.61  & 48.81 & 45.92 & 52.45 & 59.09 & 51.98 $_{(\pm0.70)}$ \\
  & & FT-M  & 40.40  & 31.46 & 27.85 & 100.00  & 27.44 & 45.43 $_{(\pm0.68)}$ & 47.03  & 49.68 & 46.99 & 100.00  & 41.41 & 57.02 $_{(\pm0.13)}$ & 48.24  & 45.55 & 43.03 & 100.00  & 23.76 & 52.12 $_{(\pm0.16)}$ \\
  & & MEND  & 15.84  & 14.35 & 17.73 & 91.74 & 70.17 & 41.97 $_{(\pm0.12)}$ & 37.22  & 38.03 & 37.19 & 91.49 & 84.10 & 57.61 $_{(\pm0.58)}$ & 6.54 & 6.51  & 6.50  & 13.52 & 20.38 & 10.69 $_{(\pm0.14)}$ \\
  & & SERAC & 83.35  & 70.80 & 80.32 & 67.66 & 13.13 & 63.05 $_{(\pm0.87)}$ & 53.58  & 45.78 & 52.42 & 56.81 & 16.90 & 45.10 $_{(\pm0.38)}$ & 43.12  & 41.69 & 38.72 & 48.08 & 14.88 & 37.30 $_{(\pm0.27)}$ \\
  & & TP  & 20.63  & 15.09 & 18.41 & 8.65  & 8.25  & 14.21 $_{(\pm0.18)}$ & 24.36  & 24.21 & 24.25 & 16.37 & 19.96 & 21.83 $_{(\pm0.14)}$ & 26.03  & 26.31 & 24.85 & 4.11  & 11.77 & 18.62 $_{(\pm0.12)}$ \\
  & & LTE & 89.32  & 82.82 & 81.51 & 94.86 & 69.83 & 83.67 $_{(\pm1.05)}$ & 61.67  & 51.05 & 61.60 & 94.78 & 90.94 & 72.01 $_{(\pm0.66)}$ & 44.52  & 43.56 & 41.29 & 96.45 & 90.86 & 63.33 $_{(\pm0.34)}$ \\
  & & RECIPE  & 84.99  & 74.20 & 82.04 & 96.82 & 87.73 & 85.16 $_{(\pm1.32)}$ & 54.64  & 46.54 & 54.10 & 94.60 & 96.93 & 69.37 $_{(\pm1.04)}$ & 43.02  & 41.63 & 38.59 & 99.68 & 92.96 & 63.18 $_{(\pm0.16)}$ \\
  & & LEMoE & 19.73  & 17.34 & 18.22 & 72.01 & 31.06 & 31.67 $_{(\pm0.14)}$ & 34.74  & 33.43 & 32.05 & 55.55 & 50.04 & 41.16 $_{(\pm0.58)}$ & 43.46  & 43.34 & 37.69 & 93.27 & 67.52 & 57.06 $_{(\pm0.38)}$ \\
  & & LiveEdit & \textbf{94.42} & \textbf{91.98}  & \textbf{84.65}  & \textbf{100.00} & \textbf{97.38}  & \textbf{93.69}  $_{(\pm0.67)}$ & \textbf{97.00} & \textbf{91.92}  & \textbf{87.53}  & \textbf{100.00} & \textbf{100.00} & \textbf{95.29}  $_{(\pm1.48)}$ & \textbf{72.86} & \textbf{70.34}  & \textbf{67.92}  & \textbf{100.00} & \textbf{100.00} & \textbf{82.22}  $_{(\pm0.62)}$\\
\bottomrule
\end{tabular}

    \caption{Lifelong editing performance on BLIP2-OPT (2.7B) and LLaVA-V1.5 (7B) across the E-VQA and VLKEB datasets.}

    \label{tab_extra_main_exp_1}
\end{table*}

\begin{table*}[!tb]
    \scriptsize
    \centering
    \setlength{\tabcolsep}{0.7pt}
    \renewcommand{\arraystretch}{0.85}

\begin{tabular}{cccccccccccccccccccc}
\toprule
\multirow{2}{*}{\textbf{\# Edit}} & \multirow{2}{*}{\textbf{Editors}} & \multicolumn{6}{c}{\textbf{E-VQA}} & \multicolumn{6}{c}{\textbf{VLKEB}} & \multicolumn{6}{c}{\textbf{E-IC}} \textbf{} \\
 & & \textbf{Rel.} & \textbf{T-Gen.} & \textbf{M-Gen.} & \textbf{T-Loc.} & \textbf{M-Loc.} & \textbf{Average} & \textbf{Rel.} & \textbf{T-Gen.} & \textbf{M-Gen.} & \textbf{T-Loc.} & \textbf{M-Loc.} & \textbf{Average} & \textbf{Rel.} & \textbf{T-Gen.} & \textbf{M-Gen.} & \textbf{T-Loc.} & \textbf{M-Loc.} & \textbf{Average} \textbf{} \\
\midrule
\multirow{9}{*}{1} & FT-L & 93.85 & 86.25 & 88.58 & 99.47 & 84.50 & 90.53 $_{(\pm0.20)}$ & 82.17 & 81.64 & 78.45 & 98.45 & 75.06 & 83.16 $_{(\pm0.99)}$ & 60.40 & 59.47 & 52.79 & 99.95 & 92.46 & 73.01 $_{(\pm1.00)}$ \\
 & FT-M & 91.42 & 85.82 & 49.09 & 100.00 & 86.72 & 82.61 $_{(\pm0.51)}$ & 93.51 & 91.28 & 68.30 & 100.00 & 81.78 & 86.97 $_{(\pm1.36)}$ & 71.60 & 67.34 & 49.55 & 100.00 & 91.69 & 76.04 $_{(\pm0.71)}$ \\
 & MEND & 93.47 & 91.32 & 91.22 & 81.51 & 74.97 & 86.50 $_{(\pm1.17)}$ & 90.28 & 89.18 & 89.55 & 93.63 & 86.69 & 89.87 $_{(\pm0.78)}$ & 90.28 & 83.91 & 79.05 & 95.92 & 93.97 & 88.63 $_{(\pm0.89)}$ \\
 & SERAC & 87.66 & 69.24 & 85.96 & 82.11 & 24.11 & 69.81 $_{(\pm0.84)}$ & 89.28 & 89.06 & 87.35 & 64.88 & 12.72 & 68.66 $_{(\pm0.76)}$ & 85.09 & 78.96 & 81.70 & 83.36 & 24.50 & 70.72 $_{(\pm0.54)}$ \\
 & TP & 52.45 & 48.39 & 51.56 & 93.93 & 83.61 & 65.99 $_{(\pm0.91)}$ & 49.17 & 51.88 & 48.92 & 90.63 & 79.79 & 64.08 $_{(\pm0.98)}$ & 52.19 & 51.49 & 49.81 & 83.33 & 72.04 & 61.77 $_{(\pm0.78)}$ \\
 & LTE & 93.54 & 77.59 & 84.07 & 94.03 & 88.14 & 87.48 $_{(\pm0.45)}$ & 92.92 & 85.16 & 86.25 & 88.29 & 81.96 & 86.91 $_{(\pm0.71)}$ & \textbf{91.30} & 84.62 & 83.55 & 94.96 & 91.88 & 89.26 $_{(\pm1.01)}$ \\
 & RECIPE & 89.23 & 70.39 & 87.94 & 95.49 & 90.90 & 86.79 $_{(\pm0.42)}$ & 93.20 & 92.40 & 90.71 & 90.29 & 84.35 & 90.19 $_{(\pm0.29)}$ & 81.25 & 74.44 & 76.96 & 97.18 & 92.17 & 84.40 $_{(\pm0.91)}$ \\
 & LEMoE & 92.59 & 90.49 & 89.16 & 97.01 & 84.33 & 90.71 $_{(\pm0.94)}$ & 92.26 & 89.28 & 88.27 & 93.87 & 67.49 & 86.23 $_{(\pm0.89)}$ & 90.32 & \textbf{86.30} & \textbf{85.75} & 96.66 & 69.17 & 85.64 $_{(\pm1.42)}$ \\
 & LiveEdit & \textbf{94.90} & \textbf{92.63} & \textbf{92.06} & \textbf{100.00} & \textbf{100.00} & \textbf{95.92} $_{(\pm0.36)}$ & \textbf{96.41} & \textbf{95.56} & \textbf{92.85} & \textbf{100.00} & \textbf{100.00} & \textbf{96.96} $_{(\pm1.04)}$ & 86.84 & 83.82 & 76.95 & \textbf{100.00} & \textbf{100.00} & \textbf{89.52} $_{(\pm1.20)}$ \\
\midrule
\multirow{9}{*}{10} & FT-L & 84.14 & 76.41 & 71.19 & 97.84 & 53.98 & 76.71 $_{(\pm0.73)}$ & 84.20 & 81.22 & 82.12 & 95.47 & 68.37 & 82.28 $_{(\pm0.73)}$ & 62.28 & 59.67 & 52.88 & 98.07 & 55.84 & 65.75 $_{(\pm0.83)}$ \\
 & FT-M & 88.05 & 81.83 & 53.26 & 100.00 & 53.01 & 75.23 $_{(\pm0.58)}$ & 91.31 & 88.82 & 63.64 & 100.00 & 63.40 & 81.43 $_{(\pm1.15)}$ & 79.04 & 70.57 & 53.82 & 100.00 & 53.19 & 71.33 $_{(\pm0.31)}$ \\
 & MEND & 37.34 & 31.66 & 35.52 & 44.31 & 37.24 & 37.21 $_{(\pm0.18)}$ & 57.01 & 55.39 & 55.34 & 72.03 & 67.66 & 61.49 $_{(\pm0.47)}$ & 69.98 & 69.35 & 59.25 & 93.39 & 91.52 & 76.69 $_{(\pm1.20)}$ \\
 & SERAC & 86.61 & 67.86 & 83.14 & 67.69 & 14.47 & 63.95 $_{(\pm0.42)}$ & 79.90 & 71.80 & 78.44 & 52.64 & 13.08 & 59.17 $_{(\pm0.41)}$ & 50.72 & 49.60 & 48.36 & 62.87 & 16.01 & 45.51 $_{(\pm0.47)}$ \\
 & TP & 35.71 & 30.29 & 36.38 & 74.53 & 55.67 & 46.52 $_{(\pm0.34)}$ & 44.70 & 47.40 & 44.74 & 69.61 & 70.01 & 55.29 $_{(\pm0.13)}$ & 47.02 & 47.29 & 46.06 & 50.10 & 42.77 & 46.65 $_{(\pm0.64)}$ \\
 & LTE & 88.55 & 73.87 & 83.91 & 89.44 & 88.02 & 84.76 $_{(\pm0.25)}$ & 89.10 & 81.76 & 85.96 & 85.53 & 81.06 & 84.68 $_{(\pm1.26)}$ & 54.77 & 55.69 & 52.24 & 90.84 & 93.06 & 69.32 $_{(\pm1.04)}$ \\
 & RECIPE & 88.44 & 69.59 & 85.61 & 90.68 & 89.78 & 84.82 $_{(\pm1.27)}$ & 82.58 & 74.01 & 81.18 & 88.33 & 85.84 & 82.39 $_{(\pm0.20)}$ & 50.38 & 48.96 & 48.11 & 92.27 & 92.69 & 66.48 $_{(\pm0.78)}$ \\
 & LEMoE & 92.55 & 86.51 & 83.31 & 79.76 & 42.61 & 76.94 $_{(\pm0.65)}$ & 89.02 & 86.94 & 85.18 & 67.60 & 50.75 & 75.90 $_{(\pm1.13)}$ & \textbf{88.13} & \textbf{84.15} & 75.86 & 92.84 & 47.92 & 77.78 $_{(\pm0.76)}$ \\
 & LiveEdit & \textbf{94.06} & \textbf{91.84} & \textbf{90.60} & \textbf{100.00} & \textbf{99.92} & \textbf{95.28} $_{(\pm0.97)}$ & \textbf{95.75} & \textbf{95.13} & \textbf{91.99} & \textbf{100.00} & \textbf{100.00} & \textbf{96.57} $_{(\pm1.50)}$ & 85.79 & 83.52 & \textbf{76.09} & \textbf{100.00} & \textbf{100.00} & \textbf{89.08} $_{(\pm1.33)}$ \\
\midrule
\multirow{9}{*}{100} & FT-L & 64.81 & 58.11 & 54.23 & 93.62 & 44.98 & 63.15 $_{(\pm1.06)}$ & 69.98 & 69.05 & 68.09 & 88.74 & 63.07 & 71.79 $_{(\pm0.16)}$ & 62.35 & 57.49 & 54.13 & 96.15 & 46.86 & 63.39 $_{(\pm0.17)}$ \\
 & FT-M & 66.06 & 51.17 & 47.24 & 100.00 & 36.65 & 60.22 $_{(\pm0.28)}$ & 60.03 & 60.18 & 57.00 & 100.00 & 51.24 & 65.69 $_{(\pm0.54)}$ & 58.93 & 54.35 & 47.84 & 100.00 & 36.80 & 59.59 $_{(\pm0.14)}$ \\
 & MEND & 23.32 & 20.75 & 21.42 & 47.27 & 42.60 & 31.07 $_{(\pm0.18)}$ & 38.85 & 40.02 & 37.99 & 72.82 & 68.89 & 51.71 $_{(\pm0.39)}$ & 51.92 & 53.96 & 48.24 & 86.02 & 83.88 & 64.80 $_{(\pm0.79)}$ \\
 & SERAC & 87.39 & 66.53 & 81.70 & 65.13 & 14.04 & 62.96 $_{(\pm0.64)}$ & 69.38 & 58.31 & 67.57 & 51.71 & 13.72 & 52.14 $_{(\pm0.78)}$ & 47.41 & 46.17 & 45.22 & 49.37 & 13.52 & 40.34 $_{(\pm0.55)}$ \\
 & TP & 19.46 & 20.32 & 22.03 & 41.85 & 32.91 & 27.32 $_{(\pm0.39)}$ & 39.60 & 42.52 & 39.14 & 36.15 & 43.68 & 40.21 $_{(\pm0.22)}$ & 39.60 & 39.64 & 38.48 & 17.29 & 26.04 & 32.21 $_{(\pm0.40)}$ \\
 & LTE & 85.05 & 71.80 & 79.61 & 87.06 & 86.10 & 81.92 $_{(\pm0.50)}$ & 79.45 & 63.55 & 79.46 & 84.18 & 79.42 & 77.21 $_{(\pm1.26)}$ & 50.47 & 49.57 & 48.17 & 88.12 & 92.87 & 65.84 $_{(\pm0.70)}$ \\
 & RECIPE & 88.26 & 67.74 & 81.48 & 91.77 & 88.23 & 83.49 $_{(\pm1.18)}$ & 71.54 & 60.04 & 70.35 & 85.57 & 86.85 & 74.87 $_{(\pm1.06)}$ & 47.27 & 46.06 & 45.12 & 89.52 & 94.25 & 64.44 $_{(\pm0.68)}$ \\
 & LEMoE & 27.84 & 26.38 & 26.73 & 68.45 & 33.73 & 36.63 $_{(\pm0.21)}$ & 44.75 & 43.33 & 41.96 & 73.90 & 60.38 & 52.86 $_{(\pm0.62)}$ & 52.85 & 50.20 & 48.11 & 92.08 & 49.63 & 58.57 $_{(\pm0.53)}$ \\
 & LiveEdit & \textbf{93.23} & \textbf{91.27} & \textbf{88.34} & \textbf{100.00} & \textbf{99.61} & \textbf{94.49} $_{(\pm0.26)}$ & \textbf{94.06} & \textbf{94.62} & \textbf{89.42} & \textbf{100.00} & \textbf{100.00} & \textbf{95.62} $_{(\pm1.26)}$ & \textbf{84.38} & \textbf{83.53} & \textbf{71.85} & \textbf{100.00} & \textbf{100.00} & \textbf{87.95} $_{(\pm0.86)}$ \\
\midrule
\multirow{9}{*}{1000} & FT-L & 58.97 & 46.34 & 50.77 & 72.04 & 41.92 & 54.01 $_{(\pm0.84)}$ & 61.85 & 60.82 & 60.41 & 76.17 & 60.34 & 63.92 $_{(\pm0.45)}$ & 58.88 & 53.98 & 52.98 & 93.31 & 46.49 & 61.13 $_{(\pm0.74)}$ \\
 & FT-M & 51.18 & 42.79 & 40.08 & 100.00 & 37.28 & 54.27 $_{(\pm0.76)}$ & 51.26 & 54.35 & 51.13 & 100.00 & 50.72 & 61.49 $_{(\pm1.03)}$ & 52.65 & 47.90 & 47.25 & 100.00 & 33.69 & 56.30 $_{(\pm0.58)}$ \\
 & MEND & 31.84 & 26.60 & 33.98 & 43.08 & 44.90 & 36.08 $_{(\pm0.15)}$ & 42.35 & 45.07 & 42.68 & 66.98 & 62.32 & 51.88 $_{(\pm0.69)}$ & 49.30 & 51.46 & 46.70 & 81.03 & 82.53 & 62.20 $_{(\pm0.30)}$ \\
 & SERAC & 84.50 & 60.36 & 81.83 & 63.26 & 13.40 & 60.67 $_{(\pm0.22)}$ & 57.02 & 51.66 & 56.80 & 48.48 & 13.24 & 45.44 $_{(\pm0.39)}$ & 47.03 & 45.79 & 44.82 & 42.12 & 15.31 & 39.01 $_{(\pm0.56)}$ \\
 & TP & 9.24 & 8.96 & 10.25 & 20.54 & 17.03 & 13.20 $_{(\pm0.19)}$ & 23.84 & 25.00 & 23.54 & 18.79 & 25.58 & 23.35 $_{(\pm0.06)}$ & 25.07 & 24.50 & 24.09 & 14.77 & 18.05 & 21.30 $_{(\pm0.30)}$ \\
 & LTE & 81.93 & 66.81 & 75.96 & 88.24 & 78.95 & 78.38 $_{(\pm0.56)}$ & 64.28 & 56.91 & 64.07 & 83.96 & 78.59 & 69.56 $_{(\pm0.51)}$ & 51.31 & 49.25 & 49.19 & 88.19 & 90.53 & 65.69 $_{(\pm0.91)}$ \\
 & RECIPE & 85.33 & 61.79 & 76.35 & 90.33 & 89.08 & 80.58 $_{(\pm1.12)}$ & 58.18 & 52.31 & 58.14 & 85.49 & 83.84 & 67.59 $_{(\pm0.29)}$ & 47.02 & 45.76 & 44.85 & 89.80 & 92.26 & 63.94 $_{(\pm0.79)}$ \\
 & LEMoE & 25.07 & 23.87 & 26.68 & 89.11 & 50.50 & 43.04 $_{(\pm0.29)}$ & 39.48 & 38.24 & 36.52 & 82.35 & 58.01 & 50.92 $_{(\pm0.12)}$ & 50.61 & 47.35 & 47.19 & 95.11 & 47.62 & 57.58 $_{(\pm0.91)}$ \\
 & LiveEdit & \textbf{92.75} & \textbf{89.81} & \textbf{85.12} & \textbf{100.00} & \textbf{96.82} & \textbf{92.90} $_{(\pm0.21)}$ & \textbf{93.41} & \textbf{89.87} & \textbf{85.21} & \textbf{100.00} & \textbf{100.00} & \textbf{93.70} $_{(\pm0.31)}$ & \textbf{81.43} & \textbf{80.34} & \textbf{62.25} & \textbf{100.00} & \textbf{99.73} & \textbf{84.75} $_{(\pm1.42)}$ \\
\bottomrule
\end{tabular}

    \caption{Lifelong editing performance on MiniGPT-4 (7B) across the E-VQA and VLKEB datasets.}
    \label{tab_extra_main_exp_2}
\end{table*}

\section{More Experiments}
\subsection{The Complete Lifelong Editing Results}
\label{appendix_complete_main_exps}
Tables \ref{tab_extra_main_exp_1} and \ref{tab_extra_main_exp_2} present the lifelong editing performance of various editors on the E-VQA \cite{MMEdit}, VLKEB \cite{MMEdit}, and E-IC \cite{MMEdit} datasets, with LLaVA \cite{llava}, BLIP2 \cite{BLIP2}, and MiniGPT-4 \cite{MiniGPT-4} as backbones, respectively. The results demonstrate a similar overall trend, confirming the effectiveness of our method. It is worth noting, however, that all methods exhibit relatively lower performance on the E-IC dataset. 
Since the image captioning task typically requires adapting descriptions of the entire image, editing involves carrying more complex information. This leads to a decline in the editing efficacy of the editors due to the limitations in representational power.

\begin{figure}[!t]
    \centering
    \includegraphics[width=1.\columnwidth]{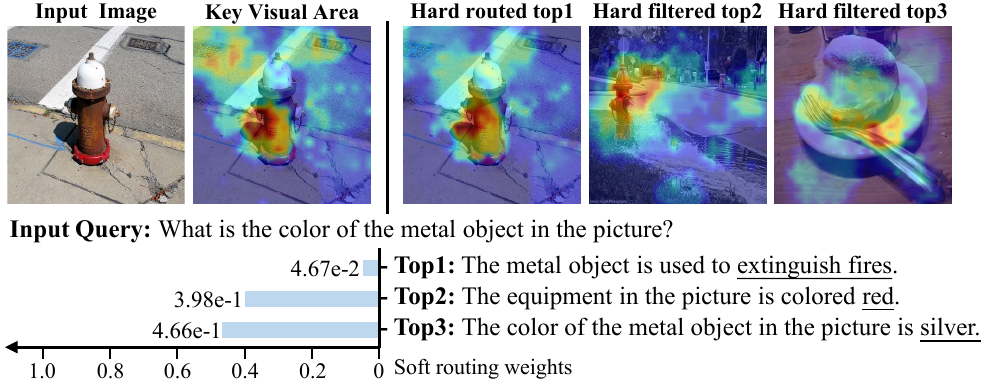}
    \caption{
     Instance analysis with difficult samples.
    } 
    \label{fig_hard_instance_analysis}
    \vspace{-1.0em}
\end{figure}

\subsection{Instance Analysis with Difficult Samples}
\label{appendix_hard_Instance_analysis}
Figure \ref{fig_hard_instance_analysis} demonstrates the necessity and complementarity of hard and soft routing. Hard routing filters out irrelevant options based on key visuals, preventing the third-ranked expert from receiving excessive weight in soft routing. Then, soft routing adjusts by reducing the weight of the top-ranked expert that was overlooked by hard routing. Therefore, relying solely on the top expert selected by hard routing can lead to errors.

\section{Other Discussions}
\noindent\textbf{Balance of Multi-Objective Losses:} The challenge in multi-goal optimization lies in balancing conflicting objectives. As shown in Fig. \ref{fig_model}, the Hard Routing Loss (HRL) doesn't share parameters with Soft Routing/Editing Loss (SRL/EL), allowing it to be optimized separately. For EL and SRL, since EL can only be minimized by selecting the correct expert in a batch, and SRL helps EL choose the right expert, they align without conflict. In EL, we align with previous work \cite{DBLP:journals/corr/abs-2408-09916}. In SRL, both losses boost semantic similarity. Consequently, we assign all loss weights as 1, and LiveEdit’s performance confirms the effectiveness of this choice.

\noindent\textbf{Time and Space Overhead of LiveEdit:} The edit time and additional memory required for each new expert remain constant, regardless of the sample size. For example, with LLaVA, it takes approximately 0.32 seconds and 160KB of memory on an A800 GPU.

\noindent\textbf{Discussion with MoVA:}
The key differences between MoVA \cite{DBLP:conf/nips/ZongMSSSJL024} and LiveEdit are as follows:
\underline{1) Task Focus:} MoVA emphasizes macro-level task performance, whereas LiveEdit focuses on the micro-level control of specific responses.
\underline{2) Intervention:} MoVA routes vision encoders externally to the LLM, while LiveEdit operates internally.
\underline{3) Routing:} In the coarse stage, MoVA uses In-Context Learning (ICL) to filter out task-irrelevant vision experts, whereas LiveEdit removes vision-irrelevant edit experts. In the fine stage, MoVA reduces single-expert bias, while LiveEdit minimizes the influence of prompt-irrelevant edit experts, complementing the previous stage.

\end{document}